%% file: ForestFireML_Main.tex
\documentclass[twoside,11pt]{article}
\usepackage[margin=0.75in]{geometry}
\usepackage{url} 

\usepackage[utf8]{inputenc}
\usepackage[english]{babel}
\usepackage{natbib}
\usepackage{hyperref}
\hypersetup{pdftex,colorlinks=true,allcolors=blue}
\usepackage{hypcap}

\usepackage{soul}
\usepackage{graphicx} 
\usepackage{xcolor}
\usepackage{imakeidx}

\usepackage{multirow}

\usepackage{amsmath}

\newcommand{\fixed}[1]{{#1}}

\newcommand{\term}[1]{\textit{#1}\index{#1}} 

\usepackage{booktabs}
\usepackage{longtable}

\usepackage{rotating}

\usepackage{authblk}

\newcommand\totalNumPub{300 }

\makeindex

\begin{document}

\title{A review of machine learning applications in wildfire science and management}

%


\author[1,2]{\textbf{Piyush Jain}\thanks{piyush.jain@canada.ca}}
\author[2]{\textbf{Sean C P Coogan}}
\author[3]{\textbf{Sriram Ganapathi Subramanian}\thanks{ss@uwaterloo.ca}}
\author[3]{\textbf{Mark Crowley} \thanks{mcrowley@uwaterloo.ca}}
\author[4]{\textbf{Steve Taylor}}
\author[2]{\textbf{Mike D Flannigan}}
\affil[1]{Natural Resources Canada,Canadian Forest Service,
Northern Forestry Centre, Edmonton, AB}
\affil[2]{Canadian Partnership for Wildland Fire Science, University of Alberta, Renewable Resources, Edmonton, AB}
\affil[3]{University of Waterloo, Electrical and Computer Engineering, Waterloo, ON}
\affil[4]{Natural Resources Canada,Canadian Forest Service,
Pacific Forestry Centre, Victoria, BC}


%
%
%
%
%
%

\maketitle

\begin{abstract}
Artificial intelligence has been applied in wildfire science and management since the 1990s, with early applications including neural networks and expert systems. Since then the field has rapidly progressed congruently with the wide adoption of machine learning (ML) methods in the environmental sciences. Here, we present a scoping review of ML applications in wildfire science and management. Our overall objective is to improve awareness of ML methods among wildfire researchers and managers, as well as illustrate the diverse and challenging range of problems in wildfire science available to ML data scientists. To that end, we first present an overview of popular ML approaches used in wildfire science to date, and then review the use of ML in wildfire science as broadly categorized into six problem domains, including: 1) fuels characterization, fire detection, and mapping; 2) fire weather and climate change; 3) fire occurrence, susceptibility, and risk; 4) fire behavior prediction; 5) fire effects; and 6) fire management. Furthermore, we discuss the advantages and limitations of various ML approaches relating to data size, computational requirements, generalizability, and interpretability, as well as identify opportunities for future advances in the science and management of wildfires within a data science context. In total, we identified \totalNumPub relevant publications \fixed{up to the end of 2019}, where the most frequently used ML methods across problem domains included random forests, MaxEnt, artificial neural networks, decision trees, support vector machines, and genetic algorithms. As such, there exists opportunities to apply more current ML methods --- including deep learning and agent based learning --- in the wildfire sciences, especially in instances involving very large multivariate datasets. We must recognize, however, that despite the ability of ML \fixed{methods} to learn on their own, expertise in wildfire science is necessary to ensure realistic modelling of fire processes across multiple scales, while the complexity of some ML methods, such as deep learning, requires a dedicated and sophisticated knowledge of their application. Finally, we stress that the wildfire research and management communities play an active role in providing relevant, high quality, and freely available wildfire data for use by practitioners of ML methods.

\end{abstract}

\textbf{Keywords:}
\textit{
machine learning, wildfire science, fire management, wildland fire, support vector machine, artificial neural network, decision trees, Bayesian networks, reinforcement learning, deep learning}

\input{introduction_section.tex}
\input{ML_section.tex}

\input{scoping_review_section.tex}
\input{applications_section.tex}
\input{discussion_section.tex}

\input{conclusions_section}

\section*{Acknowledgments}

The motivation for this paper arose from the ``Not the New Normal'' BC AI Wildfire Symposium held in Vancouver, BC, on 12 October 2018. The authors would also like to thank Intact Insurance, the Canadian Partnership for Wildland Fire Science, the NSERC Discovery Grants program and the Microsoft AI for Social Good program for their support.


\bibliographystyle{apacite}
\bibliography{ML_Fire_Review_sorted,Machine_Learning_and_wildfires_supporting_papers,ML_bibtex}

\newpage
\section*{Supplementary Material}

This supplemental contains all papers identified in this review with ML applications for wildfire science and management, organized by problem domains. Note that some papers are repeated in multiple problem domains. 

\input{ML-Fire-Paper-Table-Simple.tex}


\end{document}

%% file: introduction_section.tex
\section{Introduction}

Wildland fire is a widespread and critical element of the earth system \citep{Bond2005}, and is a continuous global feature that occurs in every month of the year. Presently, global annual area burned is estimated to be approximately 420 Mha \citep{Giglio2018}, which is greater in area than the country of India. Globally, most of the area burned by wildfires occurs in grasslands and savannas. Humans are responsible for starting over 90\% of wildland fires, and lightning is responsible for almost all of the remaining ignitions. Wildland fires can result in significant impacts to humans, either directly through loss of life and destruction to communities, or indirectly through smoke exposure. Moreover, as the climate warms we are seeing increasing impacts from wildland fire \citep{Coogan2019}. Consequently, billions of dollars are spent every year on fire management activities aimed at mitigating or preventing wildfires’ negative effects. 
Understanding and better predicting wildfires is therefore crucial in several important areas of wildfire management, including emergency response, ecosystem management, land-use planning, and climate adaptation to name a few.

Wildland fire itself is a complex process; its occurrence and behaviour are the product of several interrelated factors, including ignition source, fuel composition, weather, and topography. Furthermore, fire activity can be examined viewed across a vast range of scales,  from ignition and combustion processes that occur at a scale of centimeters over a period of seconds, to fire spread and growth over minutes to days from meters to kilometers. At larger extents, measures of fire frequency may be measured over years to millennia at regional, continental, and planetary scales (see \citet{Simard1991} for a classification of fire severity scales, and \citet{Taylor2013} for a review of numerical and statistical models that have been used to characterize and predict fire activity at a range of scales). For example, combustion and fire behavior are fundamentally physicochemical processes that can be usefully represented in mechanistic (i.e., physics-based) models at relatively fine scales \citep{Coen2018}. However, such models are often limited both by the ability to resolve relevant physical processes, as well as the quality and availability of input data \citep{Hoffman2016}. 
Moreover, with the limitations associated with currently available computing power it is not feasible to apply physical models to inform fire management and research across the larger and longer scales that are needed and in near real time. 
Thus, wildfire science and management rely heavily on the development of empirical and statistical models for meso, synoptic, strategic, and global scale processes \citep{Simard1991}, the utility of which are dependent upon their ability to represent the often complex and non-linear relationships between the variables of interest, as well as by the quality and availability of data.

	While the complexities of wildland fire often present challenges for modelling, significant advances have been made in wildfire monitoring and observation primarily due to the increasing availability and capability of remote-sensing technologies. Several satellites (eg. NASA TERRA, AQUA and GOES), for instance, have onboard fire detection sensors (e.g., Advanced Very High Resolution Radiometer (AVHRR), Moderate Resolution Imaging Spectroradiometer (MODIS), Visible Infrared Imaging Radiometer Suite (VIIRS)), and these sensors along with those on other satellites (e.g., LANDSAT series) routinely monitor vegetation distributions and changes. Additionally, improvements in numerical weather prediction  and climate models are simultaneously offering smaller spatial resolutions and longer lead forecast times \citep{Bauer2015} which potentially offer improved predictability of extreme fire weather events. \fixed{Such developments make a data-centric approach to wildfire modeling a natural evolution for many research problems given sufficient data.} Consequently, there has been a growing interest in the use of Machine Learning (ML) methodologies in wildfire science and management in recent years.  

\fixed{Although no formal definition exists, we adopt the conventional interpretation of ML as the study of computer algorithms that can improve automatically through experience \citep{Mitchell1997}.} 
This approach is necessarily data-centric, with the performance of ML algorithms dependent on the quality and quantity of available data relevant to the task at hand. 
\fixed{The field of ML has undergone an explosion of new algorithmic advances in recent years} and is deeply connected to the broader field of Artificial Intelligence (AI). 
\fixed{AI researchers aim to understand and synthesize
intelligent agents which can act appropriately to their situation and objectives, adapt to changing environments, and learn from experience \citep{Poole2010}.} 
The motivations for using AI for forested ecosystem related research, including disturbances due to wildfire, insects, and disease, were discussed in an early paper \citep{Schmoldt2001}, while \citet{Olden2008b} further argued for the use of ML methods to model complex problems in ecology. The use of ML models in the environmental sciences has seen a rapid uptake in the last decade, as is evidenced by recent reviews in the geosciences \citep{Karpatne2017}, forest ecology \citep{Liu2018}, extreme weather prediction \citep{McGovern2017}, flood forecasting \citep{Mosavi2018}, statistical downscaling \citep{Vandal2018}, remote sensing \citep{Lary2015}, and water resources \citep{Shen2018,Sun2019}. Two recent perspectives have also made compelling arguments for the application of deep learning in earth system sciences \citep{Reichstein2019} and for tackling climate change \citep{Rolnick2019}. To date, however, no such paper has synthesized the diversity of ML approaches used in the various challenges facing wildland fire science.

In this paper, we review the current state of literature on ML applications in wildfire science and management. Our overall objective is to improve awareness of ML methods among fire researchers and managers, and illustrate the diverse and challenging problems in wildfire open to data scientists. This paper is organized as follows. In Section \ref{sec:ML}, we discuss commonly used ML methods, focusing on those most commonly encountered in wildfire science. In Section \ref{sec:methods}, we give an overview of the scoping review and literature search methodology employed in this paper. In this section we also highlight the results of our literature search and examine the uptake of ML methods in wildfire science since the 1990s. In Section  \ref{sec:domains}, we review the relevant literature within six broadly categorized wildfire modeling domains: (i) Fuels characterization, fire detection, and mapping; \fixed{(ii)} fire weather and climate change; (iii) fire probability and risk; (iv) fire behavior prediction; (v) fire effects; and (vi) fire management. In Section  \ref{sec:discussion}, we discuss our findings and identify further opportunities for the application of ML methods in wildfire science and management. Finally, in Section \ref{sec:conclusions} we offer conclusions. Thus, this review will serve to guide and inform both researchers and practitioners in the wildfire community looking to use ML methods, as well as provide ML researchers the opportunity to identify possible applications in wildfire science and management.


%% file: ML_section.tex
\section{Artificial Intelligence and Machine Learning}\label{sec:ML}
\begin{quote}
    ``\textbf{Definition:  Machine Learning} - (Methods which) detect patterns in data, use the uncovered patterns to predict future data or other outcomes of interest''     
    \\ from \textit{Machine Learning: A Probabilistic Perspective, 2012} \citep{Murphy2012}.
\end{quote}

ML itself can be seen as a branch of AI or statistics, depending who you ask, that focuses on building predictive, descriptive, or actionable models for a given problem by using collected data\fixed{, or incoming data,} specific to that problem. ML methods learn directly from data and dispense with the need for a large number of expert rules or the need to model individual environmental variables \fixed{with perfect accuracy}. ML algorithms develop their own internal model of the underlying \fixed{distributions} when learning from data and thus need not be explicitly provided with physical properties of different parameters. \fixed{Take for example, the task of modeling wildland fire spread, the relevant physical properties which include fuel composition, local weather and topography. The} current state of the art in wildfire prediction includes physics-based simulators that fire fighters and strategic planners rely on to take many critical decisions regarding allocation of scarce firefighting resources in the event of a wildfire \citep{sullivan2007review}. These physics-based simulators, however, have certain critical limitations; they normally render very low accuracies, have a prediction bias in regions where they are designed to be used, are often hard to design and implement due to the requirement of large number of expert rules. Furthermore, modelling many complex environmental variables is often difficult due to large resource requirements and complex or heterogeneous data formats. ML algorithms, however, learn their own mappings between parametric rules directly from data and do not require expert rules, which is particularly advantageous when the number of parameters are quite large and their physical properties quite complex, as in the case of wildland fire. Therefore, a ML approach to wildfire response may help to avoid many of the limitations of physics-based simulators. 

\begin{figure}[t]
\begin{center}
\includegraphics[width=0.9\linewidth]{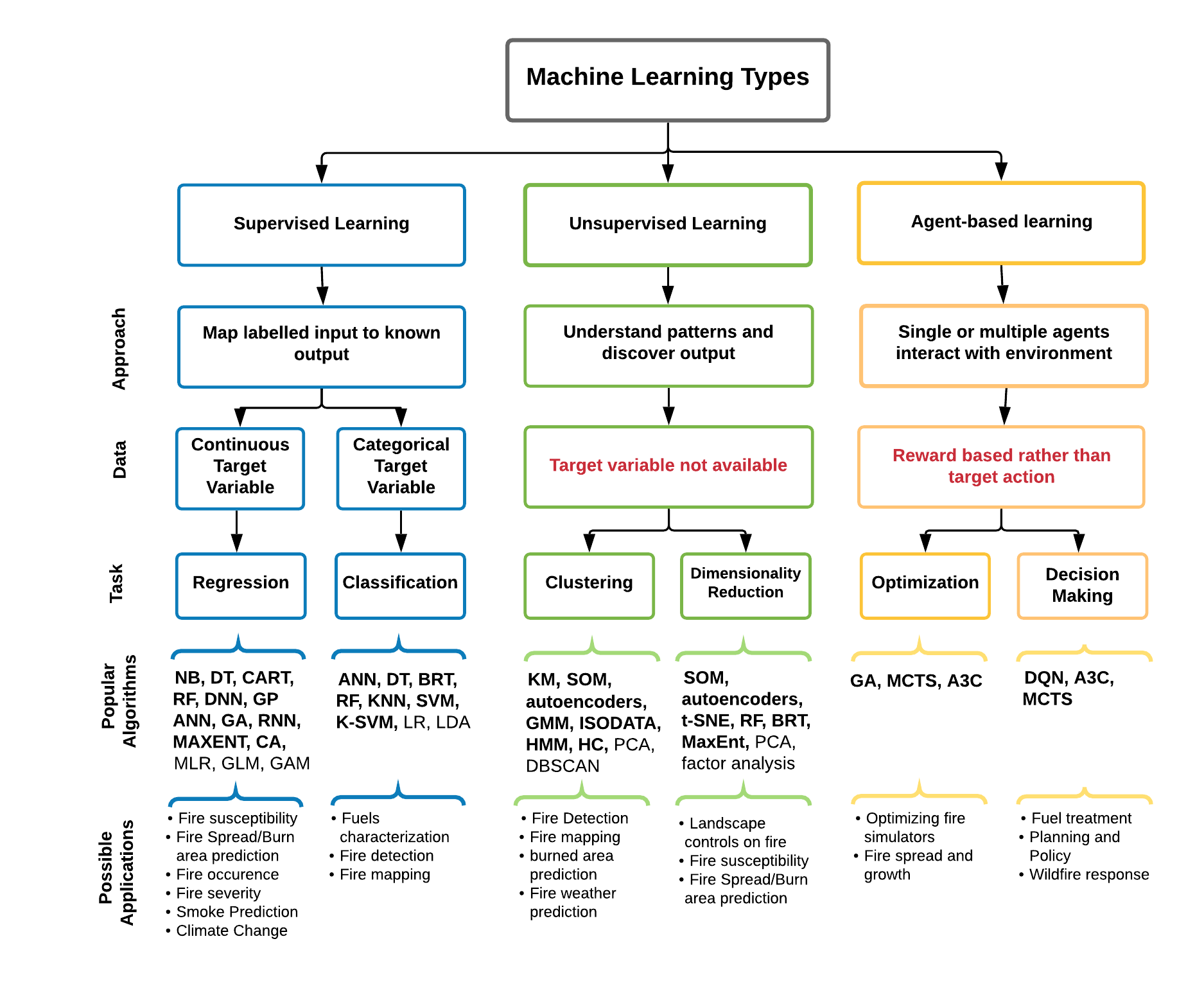}
\caption{A diagram showing the main machine learning types, types of data, and modeling tasks in relation to popular algorithms and potential applications in wildfire science and management. Note that the algorithms shown bolded are core ML methods whereas non-bolded algorithms are often not considered ML. 
}
\label{fig:AIMLimage}
\end{center}
\end{figure}

	A major goal of this review is to provide an overview of the various ML methods utilized in wildfire science and management. Importantly, we also provide a generalized framework for guiding wildfire scientists interested in applying ML methods to specific problem domains in wildland fire research. This conceptual framework, derived from the approach in \citep{Murphy2012} and modified to show examples relevant to wildland fire and management is shown in Fig. \ref{fig:AIMLimage}. In general, ML methods can be identified as \fixed{belonging} to one of three types: supervised learning; unsupervised learning; or, agent based learning. We describe each of these below.

	\textbf{Supervised Learning} - In supervised ML all problems can be seen as one of learning a parametrized function, often called a ``model'', that maps inputs (i.e., predictor variables) to outputs (or “target variables”) both of which are known. The goal of supervised learning is to use an algorithm to learn the parameters of that function using available data. In fact, both linear and logistic regression can be seen as very simple forms of supervised learning. Most of the successful and popular ML methods fall into this category.

	\textbf{Unsupervised Learning} - If the target variables are not available, then ML problems are typically much harder to solve. In unsupervised learning, the canonical tasks are dimensionality reduction and clustering, where relationships or patterns are extracted from the data without any guidance as to the ``right'' answer. Extracting embedded dimensions which minimize variance, or assigning datapoints to (labelled) classes which maximize some notion of natural proximity or other measures of similarity are examples of unsupervised ML tasks.

	\textbf{Agent Based Learning} - Between supervised and unsupervised learning are a group of ML methods where learning happens by simulating behaviors and interactions of a single or a group of autonomous agents. These are general unsupervised methods which use incomplete information about the target variables, (i.e., information is available for some instances but not others), requiring generalizable models to be learned. A specific case in this space is Reinforcement Learning~\citep{sutton1998introduction}, which is used to model decision making problems over time where critical parts of the environment can only be observed interactively through trial and error. This class of problems arises often in the real world and require efficient learning and careful definition of values (or preferences) and exploration strategies.

	In the next section, we present a brief introduction to commonly used ML methods from the aforementioned learning paradigms. We note that this list is not meant to be exhaustive, and that some methods can accommodate both supervised and unsupervised learning tasks.
It should be noted that the classification of a method as belonging to either ML or traditional statistics is often a question of taste. For the purpose of this review --- and in the interests of economy --- we have designated a number of methods as belonging to traditional statistics rather than ML. For a complete listing see tables 1 and 2.

\input{ML_acronyms_table.tex}

\subsection{Decision Trees}

Decision Trees (DT) \citep{breiman2017classification} belong to the class of supervised learning algorithms and are another example of a universal function approximator, although in their basic form such universality is difficult to achieve. DTs can be used for both classification and regression problems. A decision tree is a set of if-then-else rules with multiple branches joined by decision nodes and terminated by leaf nodes. The decision node is where the tree splits into different branches, with each branch corresponding to the particular decision being taken by the algorithm whereas leaf nodes represent the model output. This could be a label for a classification problem or a continuous value in case of a regression problem. A large set of decision nodes is used in this way to build the DT. The objective of DTs are to accurately capture the relationships between input and outputs using the smallest possible tree that avoids overfitting. C4.5 \citep{quinlan1993} and Classification and Regression Trees (CART, \citep{breiman1984}) are examples of common single DT algorithms. Note that while the term CART is also used as an umbrella term for single tree methods, we use DT here to refer to all such methods. The majority of decision tree applications are ensemble decision tree (EDT) models that use multiple trees in parallel (ie. bootstrap aggregation or bagging) or sequentially (ie., boosting) to arrive at a final model. In this way, EDTs make use of many weak learners to form a strong learner while being robust to overfitting. EDTs are well described in many ML/AI textbooks and are widely available as implemented libraries.

\subsubsection{Random Forests}

A Random Forest (RF) \citep{Breiman2001} is an ensemble model composed of a many individually trained DTs, and is the most popular implementation of a bagged decision tree. Each component DT in a RF model makes a classification decision where the class with the maximum number of votes is determined to be the final classification for the input data. RFs can also be used for regression where the final output is determined by averaging over the individual tree outputs. The underlying principle of the RF algorithm is that a random subset of features is selected at each node of each tree; the samples for training each component tree are selected using bagging, which resamples (with replacement) the original set of datapoints. The high performance of this algorithm is achieved by minimizing correlation between trees while reducing model variance so that a large number of different trees provides greater accuracy than individual trees. However, this improved performance comes at the cost of an increase in bias and loss of interpretability (although variable importance can still be inferred through permutation tests).

\subsubsection{Boosted Ensembles}

Boosting describes a strategy where one combines a set of weak learners --- usually decision trees --- to make a strong learner using a sequential additive model. Each successive model improves on the previous by taking into account the model errors from the previous model, which can be done in more than one way. For example, the adaptive boosting algorithm, known as AdaBoost \citep{freund1995cc}, works by increasing the weight of observations that were previously misclassified. This can in principle reduce the classification error leading to a high level of precision \citep{Hastie2009}.

Another very popular implementation for ensemble boosted trees is Gradient Boosting Machine (GBMs), which makes use of the fact that each DT model represents a function that can be differentiated with respect to its parameters, i.e., how much a change in the parameters will change the output of the function. GBMs sequentially build an ensemble of multiple weak learners by following a simple gradient which points in the opposite direction to weakest results of the current combined model \citep{friedman2001}. 

The details for the GBM algorithm are as follows. Denoting the target output as $Y$, and given a tree-based ensemble model, represented as a function $T_i(X)\rightarrow Y$, after adding $i$ weak learners already, the ``perfect'' function for the $(i+1){th}$ weak learner would be $h(x)=T_i(x) - Y$ which exactly corrects the previous model (i.e., $T_{(i+1)}(x) = T_i(x) + h(x) = Y$).  In practice, we can only approach this perfect update by performing functional gradient descent where we use an approximation of the true residual (i.e., loss function) at each step. In our case this approximation is simply the sum of the residuals from each weak learner decision tree $L(Y, T(X)) = \sum_i Y-T_i(X)$.
GBM explicitly uses the gradient $\nabla_{T_i} L(Y,T_i(X)$ of the loss function of each tree to fit a new tree and add it to the ensemble. 

In a number of domains, and particularly in the context of ecological modeling GBM is often referred to as Boosted Regression Trees (BRTs) \citep{Elith2008}. For consistency with the majority of literature reviewed in this paper we henceforth use the latter term.
It should be noted that while deep neural networks (DNNs) and EDT methods are both universal function approximators, EDTs are more easily interpretable and faster to learn with less data than DNNs. However, there are fewer and fewer cases where trees-based methods can be shown to provide superior performance on any particular metric when DNNs are trained properly with enough data (see for example, \citet{Korotcov2017}).

\subsection{Support Vector Machines}

Another category of supervised learning includes Support Vector Machines (SVM) \citep{hearst1998support} and related kernel-based methods. SVM is a classifier that determines the hyper-plane (decision boundary) in an n-dimensional space separating the boundary of each class, for data in n dimensions. SVM finds the optimal hyper-plane in such a way that the distance between the nearest point of each class to the decision boundary is maximized. If the data can be separated by a line then the hyper-plane is defined to be of the form $w^Tx + b = 0$ where the $w$ is the weight vector, $x$ is the input vector and $b$ is the bias. The distance of the hyper-plane to the closest data point $d$, called a support vector, is defined as the margin of separation. The objective is to find the optimal hyper-plane that minimizes the margin. If they are not linearly separable, kernel SVM methods such as Radial Basis Functions (RBF) first apply a set of transformations to the data to a higher dimensional space where finding this hyperplane would be easier. 
SVMs have been widely used for both classification and regression problems, although recently developed deep learning algorithms have proved to be more efficient than SVMs given a large amount of training data. However, for problems with limited training samples, SVMs might give better performances than deep learning based classifiers.

\subsection{Artificial Neural Networks and Deep Learning}
	
The basic unit of an Artificial Neural Network (ANN) is a neuron (also called a perceptron or logistic unit). A neuron is inspired by the functioning of neurons in mammalian brains in that it can learn simple associations, but in reality it is much simpler than its biological counterpart. A neuron has a set of inputs which are combined linearly through multiplication with weights associated with the input. The final weighted sum forms the output signal which is then passed through a (generally) non-linear activation function. Examples of activation functions include sigmoid, tanh, and the Rectified Linear Unit (ReLU). This non-linearity is important for general learning since it creates an abrupt cutoff (or threshold) between positive and negative signals. The weights on each connection represent the function parameters which are fit using supervised learning by optimizing the threshold so that it reaches a maximally distinguishing value. 

\begin{figure}[t]
\begin{center}
\includegraphics[width=.7\linewidth]{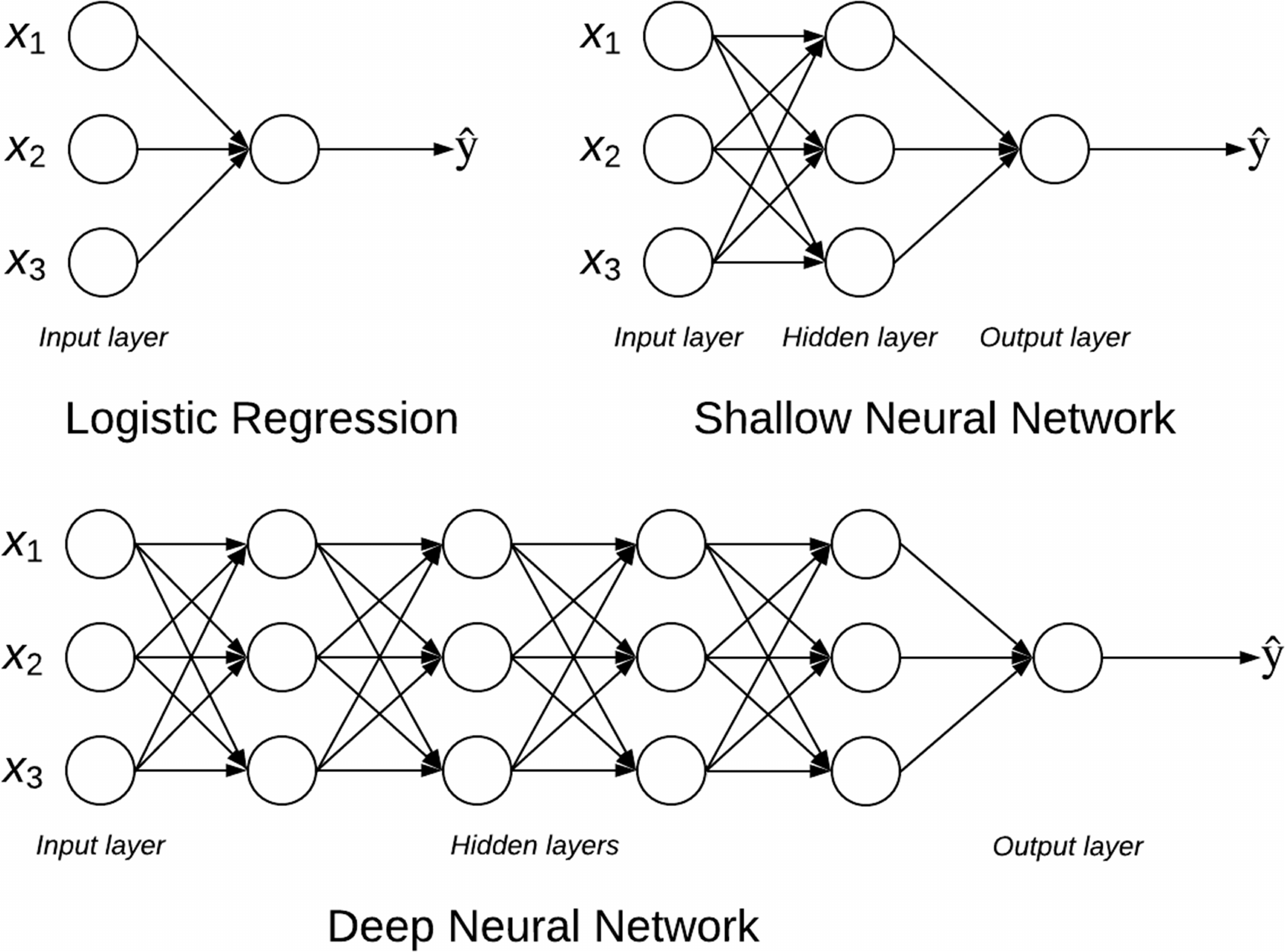}
\caption{Logistic regression can be seen as basic building block for neural networks,  with no hidden layer and a sigmoid activation function. Classic shallow neural networks (also known as Multi-Layer Perceptrons) have at least one hidden layer and can have a variety of activation functions. Deep neural networks essentially have a much larger number of hidden layers as well as use additional regularization and optimization methods to enhance training.}
\label{fig:DeepNetworks}
\end{center}
\end{figure}

	In practice, even simple ANNs, often called Multi-Layered Perceptrons (MLP), combine many neuron units in parallel, each processing the same input with independent weights. In addition, a second layer of hidden neuron units can be added to allow more degrees of freedom to fit general functions, see Figure \ref{fig:DeepNetworks}(a). MLPs are capable of solving simple classification and regression problems. For instance, if the task is one of classification, then the output is the predicted class for the input data, whereas in the case of a regression task the output is the regressed value for the input data. Deep learning \citep{lecun2015deep} refers to using Deep Neural Networks (DNNs) which are ANNs with multiple hidden layers (nominally more than 3) and include Convolutional Neural Networks (CNNs) popularized in image analysis and Recurrent Neural Networks (RNNs) which can be used to model dynamic temporal phenomena. The architecture of DNNs can vary in connectivity between nodes, the number of layers employed, the types of activation functions used, and many other types of hyperparameters. Nodes within a single layer can be fully connected, or connected with some form of convolutional layer (e.g., CNNs), recurrent units (e.g., RNNs), or other sparse connectivity. The only requirement of all these connectivity structures and activation functions is that they are differentiable.

Regardless of the architecture, the most common process of training a ANN involves processing input data fed through the network layers and activation functions to produce an output. In the supervised setting, this output is then compared to the known true output (i.e., labelled training data) resulting in an error measurement (loss or cost function) used to evaluate model performance. The error for DNNs are commonly calculated as a cross entropy loss between the predicted output label and the true output label. Since every part of the network is mathematically differentiable we can compute a gradient for the entire network. This gradient is used to calculate the proportional change in each network weight needed to produce an infinitesimal increase in the likelihood of the network producing the same output for the most recent output. The gradient is then weighted by the computed error, and thereafter all the weights are updated in sequence using a backpropagation algorithm \citep{hecht1992theory}.

ANNs can also be configured for unsupervised learning tasks. For example, self-organizing maps (SOMs) are a form of ANN adapted for dealing with spatial data and have therefore found widespread use in the atmospheric sciences \citep{skific2012self}. A SOM is a form of unsupervised learning that consists of a two-dimensional array of nodes as the input layer, representing say, a gridded atmospheric variable at a single time. The algorithm clusters similar atmospheric patterns together and results in a dimensionality reduction of the input data. More recently, unsupervised learning methods from deep learning, such as autoencoder networks, are starting to replace SOMs in the environmental sciences \citep{Shen2018}.
	
\subsection{Bayesian methods}

\subsubsection{Bayesian Networks}

Bayesian networks (Bayes net, belief network; BN) are a popular tool in many applied domains because they provide an intuitive graphical language for specifying the probabilistic relationships between variables as well as tools for calculating the resulting probabilities \citep{pearl88}. The basis of BNs is Bayes’ theorem, which relates the conditional and marginal probabilities of random variables. BNs can be treated as a ML task if one is trying to automatically fit the parameters of the model from data, or even more challenging, to learn the best graphical structure that should be used to represent a dataset. BNs have close ties to causal reasoning, but it is important to remember that the relationships encoded in a BN are inherently correlational rather than causal.
	BNs are acyclic graphs, consisting of nodes and arrows (or arcs), defining a probability distribution over variables $\mathcal{U}$.  The set of parents of a node (variable) $X$, denoted $\pi_X$, are all nodes with directed arcs going into $X$. BNs provide compact representation of conditional distributions since $p(X_i|X_1,\ldots,X_{i-1}) = p(X_i|\pi_{X_i})$ where $X_1,\ldots,X_{i-1}$ are arranged to be all of the ancestors of $X_i$ other than its direct parents. Each node $X$ is associated with a conditional probability table over $X$ and its parents defining $p(X|\pi_X)$. If a node has no parents, a prior distribution is specified for $p(X)$. The joint probability distribution of the network is then specified by the chain rule $P(U) = \prod_{X\in \mathcal{U}} p(X|\pi_X)$.

\subsubsection{Naïve Bayes}

A special case of a BN is the Naïve Bayes (NB) classifier, which assumes conditional independence between input features, which allows the likelihood function to be constructed by a simple multiplication of the conditional probability of each input variable conditional on the output. Therefore, while NB is fast and straightforward to implement, prediction accuracy can be low for problems where the assumption of conditional independence does not hold.

\subsubsection{Maximum Entropy}

Maximum Entropy (MaxEnt), originally introduced by \citet{Phillips2006}, is a presence only framework that fits a spatial probability distribution by maximising entropy, consistent with existing knowledge. MaxEnt can be considered a Bayesian method since it is compatible with an application of Bayes Theorem as existing knowledge is equivalent to specifying a prior distribution. MaxEnt has found widespread use in landscape ecology species distribution modeling \citep*{Elith2011}, where prior knowledge consists of occurrence observations for the species of interest.

\subsection{Reward based methods}

\subsubsection{Genetic Algorithms}

Genetic algorithms (GA) are heuristic algorithms inspired by Darwin's theory of evolution (natural selection) and belong to a more general class of evolutionary algorithms \citep{Mitchell1996}. GAs are often used to generate solutions to search and optimization problems by using biologically motivated operators such as mutation, crossover, and selection. In general, GAs involve several steps. The first step involves creating an initial population of potential solutions, with each solution encoded as a chromosome. Second a fitness function appropriate to the problem is defined, which returns a fitness score determining how likely an individual is to be chosen for reproduction. The third step requires the selection of pairs of individuals, denoted as parents. In the fourth step, a new population of finite individuals are created by generating two new offspring from each set of parents using crossover, whereby a new chromosome is created by some random selection process from each parents chromosomes. In the final step called mutation, a small sample of the new population is chosen and a small perturbation is made to the parameters to maintain diversity. The entire process is repeated many times until the desired results are satisfactory (based on the fitness function), or some measure of convergence is reached.

\subsubsection{Reinforcement Learning}

Reinforcement learning (RL) represents a very different learning paradigm to supervised or unsupervised learning. In RL, an agent (or actor) interacts with its environment and learns a desired behavior (set of actions) in order to maximize some reward. RL is a solution to a Markov Decision Process (MDP) where the transition probabilities are not explicitly known but need to be learned. This type of learning is well suited to problems of automated decision making, such as required for automated control (e.g., robotics) or for system optimization (e.g., management policies). Various RL algorithms include Monte Carlo Tree Search (MTCS), Q-Learning, and Actor-Critic algorithms. For an introduction to RL see \citet{Sutton2018}.

\subsection{Clustering methods}

Clustering is the process of splitting a set of points into groups where each point in a group is more similar to its own group than any other group. There are different ways in which clustering can be done, for example, the K-means (KM) clustering algorithm \citep{macqueen1967some}, based on a centroid model, is perhaps the most well-known clustering algorithm. In K-means, the notion of similarity is based on closeness to the centroid of each cluster. K-means is an iterative process in which the centroid of a group and points belonging to a group are updated at each step. The K-means algorithm consists of five steps: (i) specify the number of clusters; (ii) each data point is randomly assigned to a cluster; (iii) the centroids of each cluster is calculated; (iv) the points are reassigned to the nearest centroids, and (v) cluster centroids are recomputed. Steps iv and v repeat until no further changes are possible. Although KM is the most widely used clustering algorithm, several other clustering algorithms exist including, for example, agglomerative Hierarchical Clustering (HC), Gaussian Mixture Models (GMMs) and Iterative Self-Organizing DATA (ISODATA).

\subsection{Other methods}

\subsubsection{K-Nearest Neighbor}
The K-Nearest Neighbors (KNN) algorithm is a simple but very effective supervised classification algorithm which is based on the intuitive premise that similar data points are in close proximity according to some metric \citep{Altman1992}. Specifically, a KNN calculates the similarity of data points to each other using the Euclidean distance between the $K$ nearest data points. The optimal value of $K$ can be found experimentally over a range values using the classification error. 
\fixed{KNN is widely used in applications where a search query is performed such that results should be similar to another pre-existing entity. Examples of this include finding similar images to a specified image and recommender systems. Another popular application of KNN is outlier (or anomaly) detection, whereby the points (in a multidimensional space) farthest away from their nearest neighbours may be classified as outliers.}

\subsubsection{Neuro-Fuzzy models}
	
Fuzzy logic is an approach for encoding expert human knowledge into a system by defining logical rules about how different classes overlap and interact without being constrained to ``all-or-nothing’’ notions of set inclusion or probability of occurrence. Although early implementations of fuzzy logic systems depended on setting rules manually, and therefore are not considered machine learning, using fuzzy rules as inputs or extracting them from ML methods are often described as ``neuro-fuzzy'' methods. For example, the Adaptive Neuro-Fuzzy Inference System (ANFIS) \citep{jang1993} fuses fuzzy logical rules with an ANN approach, while trying to maintain the benefits of both. ANFIS is a universal function approximator like ANNs. However, since this algorithm originated in the 1990s, it precedes the recent deep learning revolution so is not necessarily appropriate for very large data problems with complex patterns arising in high-dimensional spaces. Alternatively, human acquired fuzzy rules can be integrated into ANNs learning; however, it is not guaranteed that the resulting trained neural network will still be interpretable. It should be noted that fuzzy rules and fuzzy logic are not a major direction of research within the core ML community.

%% file: ML_acronyms_table.tex
\begin{table}[t!]
\begin{tabular}{lll}
\hline
\multicolumn{2}{l}{Machine Learning Methods} & \\
\hline
A3C & Asynchronous Advantage Actor-Critic  & \\
AdaBoost & Adaptive Boosting & \\
ANFIS & Adaptive Neuro Fuzzy Inference System & \\
ANN & Artificial Neural Networks &  \\
ADP & Approximate Dynamic Programming (a.k.a. reinforcement learning) & \\
Bag & Bagged Decision Trees & \\
BN  & Bayesian Networks          &  \\
BRT & Boosted Regression Trees (a.k.a. Gradient Boosted Machine) & \\
BULC & Bayesian Updating of Land Cover & \\
CART & Classification and Regression Tree & \\
CNN & Convolutional Neural Network & \\
DNN & Deep Neural Network & \\
DQN & Deep Q-Network & \\
DT  & Decision Trees (incl. CART, J48, jRip)            &  \\
EDT & Ensemble Decision Trees (incl. bagging and boosting) & \\
ELM & Extreme Machine Learning (i.e., feedforward network) & \\
GA  & Genetic algorithms (a.k.a evolutionary algorithms)        &  \\
GBM & Gradient Boosted Machine (a.k.a. Boosted Regression Trees, incl. XGBoost, AdaBoost, LogitBoost) &\\
GMM & Gaussian Mixture Models & \\
GP  & Gaussian Processes         &  \\

HCL & Hard Competitive Learning  & \\
HMM & Hidden Markov Models & \\
ISODATA & Iterative Self-Organizing DATA algorithm & \\
KNN & K Nearest Neighbor         &  \\
KM  & K-means Clustering         &  \\
LB  & LogitBoost (incl. AdaBoost) &  \\
LSTM & Long Short Term Memory & \\
MaxEnt & Maximum Entropy & \\
MCMC & Markov Chain Monte Carlo & \\
MCTS & Monte Carlo Tree Search & \\
MLP & Multilayer Perceptron & \\
MDP & Markov Decision Process & \\
NB & Naive Bayes & \\
NFM & Neuro-Fuzzy models & \\
PSO & Particle Swarm Optimization & \\
RF  & Random Forest              &  \\
RL  & Reinforcement Learning     &  \\
RNN & Recurrent Neural Network & \\
SGB & Stochastic Gradient Boosting & \\
SOM & Self-organizing Maps & \\
SVM & Support Vector Machines    & \\
t-SNE & T-distributed Stochastic Neighbor Embedding & \\
\bottomrule
\end{tabular}
\caption{Table of acronyms and definitions for common machine learning algorithms referred to in text.}
\label{tab:acro}
\end{table}

\begin{table}[t!]
\begin{tabular}{lll}
\hline
\multicolumn{2}{l}{Non-machine learning methods} & \\
\hline
DBSCAN & Density-based spatial clustering of applications with noise & \\
GAM & Generalized Additive Model & \\
GLM & Generalized Linear Model & \\
KLR & Kernel Logistic Regression & \\
LDA & Linear Discriminant Analysis & \\
LR & Logistic Regression & \\
MARS & Multivariate Adaptive Regression Splines & \\
MLR & Multiple Linear Regression & \\
PCA & Principal Component Analysis & \\
SLR & Simple Linear regression & \\
\bottomrule
\end{tabular}
\caption{Table of acronyms and definitions for common data analysis algorithms usually considered as foundational to, or outside of, machine learning itself.}
\label{tab:acro2}
\end{table}

%% file: scoping_review_section.tex
\section{Literature search and scoping review}\label{sec:methods}

The combination of ML and wildfire science and management comprises a diverse range of topics in a relatively nascent field of multidisciplinary research. Thus, we employed a scoping review methodology \citep{Arksey2005,Levac2010} for this paper. The goal of a scoping review is to characterize the existing literature in a particular field of study, particularly when a topic has yet to be extensively reviewed and the related concepts are complex and heterogeneous \citep*{Pham2014}. Furthermore, scoping reviews can be particularly useful for summarizing and disseminating research findings, and for identifying research gaps in the published literature. A critical review of methodological advances and limitations and comparison with other methods is left for future work. We performed a literature search using the Google Scholar and Scopus databases and the key words  ``wildfire'' or ``wildland fire'' or``forest fire'' or ``bushfire'' in combination with ``machine learning'' or ``random forest'' or ``decision trees'' or ``regression trees'' or ``support vector machine'' or ``maximum entropy'' or ``neural network'' or ``deep learning'' or ``reinforcement learning''. We also used the Fire Research Institute online database (\url{http://fireresearchinstitute.org}) using the following search terms: ``Artificial Intelligence''; ``Machine Learning''; ``Random Forests''; ``Expert Systems''; and ``Support Vector Machines''. Furthermore, we obtained papers from references cited within papers we had obtained using literature databases.

\begin{figure}[t!]
    \centering
\includegraphics[scale=0.8]{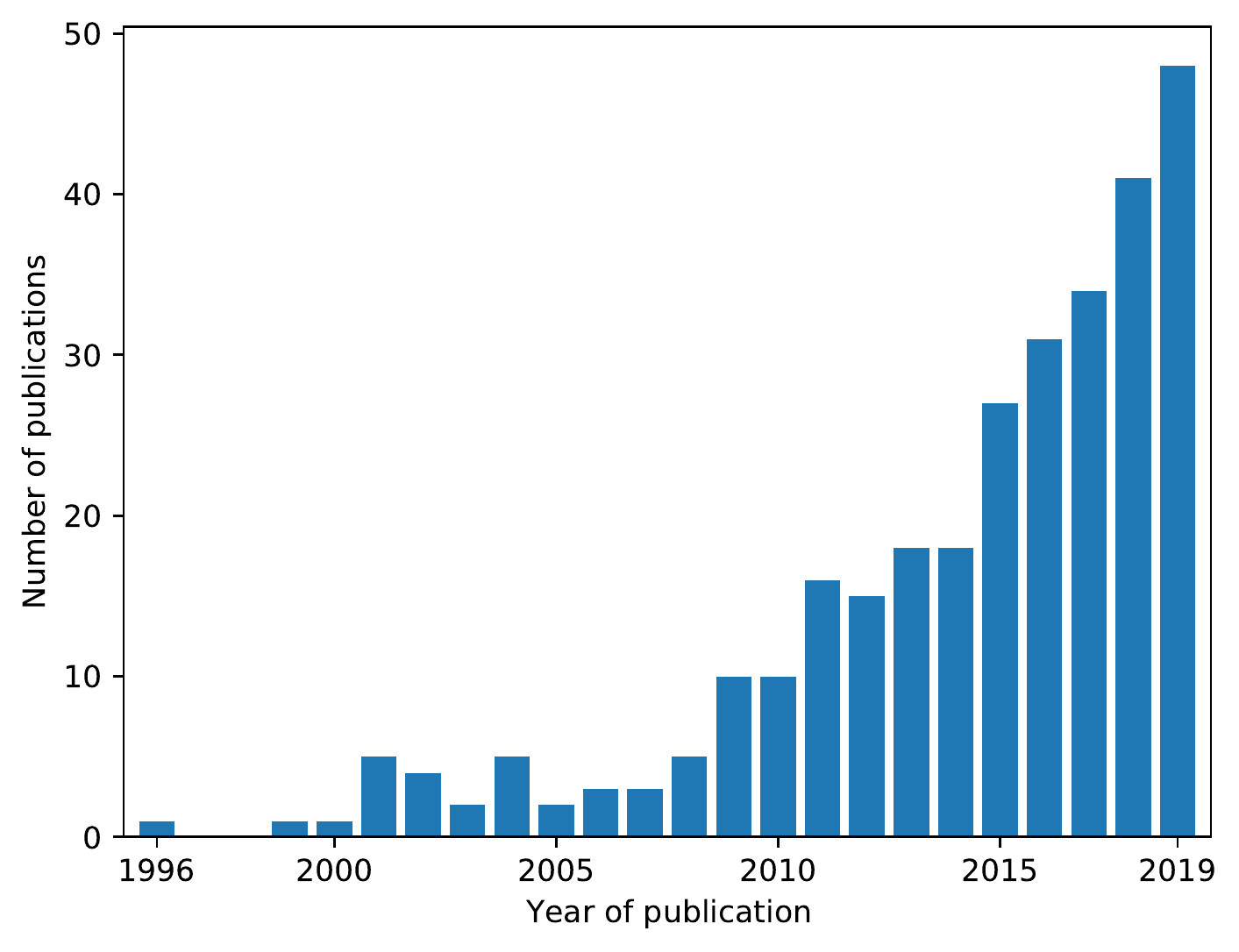} 
\caption{Number of publications by year for \totalNumPub publications on topic of ML and wildfire science and management as identified in this review.}
\label{fig:articles_review}
\end{figure}

\begin{figure}[t!]
    \centering
\includegraphics[scale=0.8]{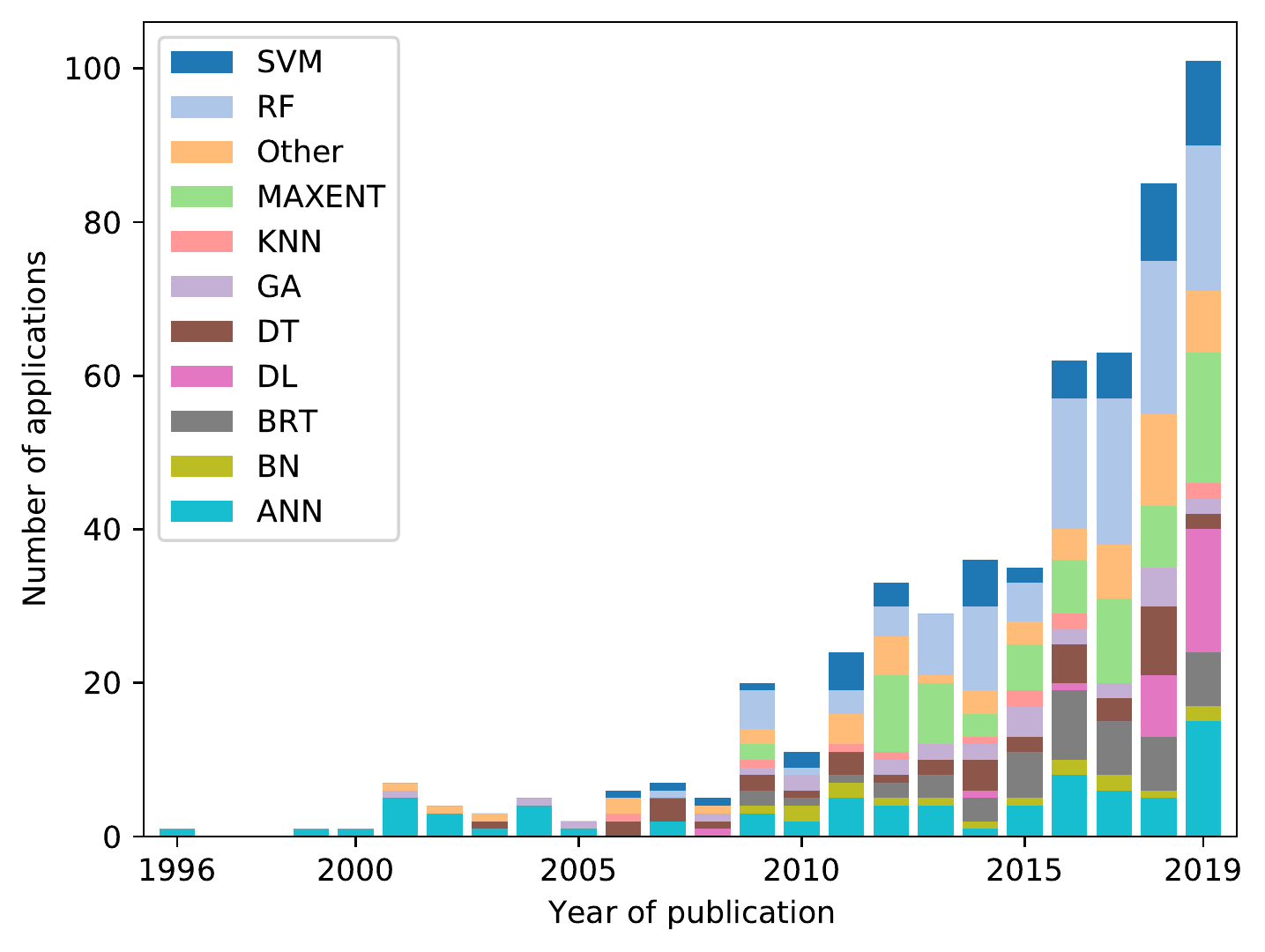} 
\caption{Number of ML applications by category and by year for \totalNumPub publications on topic of ML and wildfire science and management as identified in this review.}
\label{fig:articles_review2}
\end{figure}

After performing our literature search, we identified a total of \ \totalNumPub publications relevant to the topic of ML applications in wildfire science and management (see supplementary material for a full bibliography). Furthermore, a search of the Scopus database revealed a dramatic increase in the number of wildfire and ML articles published in recent years (see Fig. \ref{fig:articles_review}). After identifying publications for review, we further applied the following criteria to exclude non-relevant or unsuitable publications, including: (i) conference submissions where a journal publication describing the same work was available; (ii) conference posters; (iii) articles in which the methodology and results were not adequately described to conduct an assessment of the study; (iv) articles not available to as either by open access or by subscription; and (v) studies that did not present new methodologies or results. 

%% file: applications_section.tex
\section{Wildfire applications}\label{sec:domains}

In summary, we found a total of \totalNumPub journal papers or conference proceedings on the topic of ML applications in wildfire science and management. We found the problem domains with the highest application of ML methods was \term{Fire Occurrence}, \term{Susceptibility and Risk} (127 papers) followed by \term{Fuels Characterization}, \term{Fire Detection And Mapping} (\fixed{66} papers), \term{Fire Behaviour Prediction} (43 papers), \term{Fire Effects} (\fixed{35} papers), \term{Fire Weather and Climate Change} (20 papers), and \term{Fire Management} (16 papers). Within Fire Occurrence, Susceptibility and Risk, the subdomains with the most papers were \term{Fire Susceptibility Mapping} (71 papers) and \term{Landscape Controls on Fire} (101 papers). Refer to \fixed{table \ref{tab:summary} and the supplementary material} for a break-down of each problem subdomain and ML methods used, as well as study areas considered. 

\begin{sidewaystable}
  \centering
\input{subdomains_methods_count_nice_gt5.tex}
  \caption{Summary of application of ML methods applied to different problem domains in wildfire science and management. A table of acronyms for the ML methods are given in \ref{tab:acro}.  Note that in some cases a paper may use more than one ML method and/or appear in multiple problem domains.}
  \label{tab:summary}
\end{sidewaystable}

\subsection{Fuels Characterization, Fire Detection, and Mapping}

\subsubsection{Fuels characterization}
\index{Fuels Characterization}

    
Fires ignite in a few fuel particles; subsequent heat transfer between particles through conduction, radiation and convection, and the resulting fire behavior (fuel consumption, spread rate, intensity) is influenced by properties of the live and dead vegetative fuels, including moisture content, biomass, and vertical and horizontal distribution.  Fuel properties are a required input in all fire behavior models, whether it be a simple categorical vegetation type, as in the Canadian FBP System, or as physical quantities in a 3 dimensional space (eg. see FIRETEC model). Research to predict fuel properties has been carried out at two different scales 1) regression applications to predict quantities such as the crown biomass of single trees from more easily measured variables such as height and diameter, and 2) classification applications to map fuel type descriptors or fuel quantities over a landscape from visual interpretation of air photographs or by interpretation of the spectral properties of remote sensing imagery. However, relatively few studies have employed ML to wildfire fuel prediction, leaving the potential for substantially more research in this area.

In an early study, \citet{Riano2005} used an ANN to predict and map the equivalent water thickness and dry matter content of wet and dry leaf samples from 49 species of broad leaf plants using reflectance and transmittance values in the Ispra region of Italy. \citet{Pierce2012} used RF to classify important canopy fuel variables (e.g. canopy cover, canopy height, canopy base height, and canopy bulk density) related to wildland fire in Lassen Volcanic National Park, California, using field measurements, topographic data, and NDVI to produce forest canopy fuel maps. Likewise, \citet{Riley2014} used RF with Landfire and biophysical variables to perform fuel classification and mapping in Eastern Oregon. The authors of the aforementioned study achieved relatively high overall modelling accuracy, for example, 97\% for forest height, 86\% for forest cover, and 84\% for existing vegetation group (i.e. fuel type). \citet{Lopez-Serrano2016} compared the performance of three common ML methods (i. SVM; ii. KNN; and iii. RF) and \term{multiple linear regression} in estimating above ground biomass in the Sierra Madre Occidental, Mexico. The authors reported the advantages and limitations of each method, concluding that that the \term{non-parametric} ML methods had an advantage over multiple linear regression for biomass estimation. \citet{Garcia2011} used SVM to classify LiDAR and multispectral data to map fuel types in Spain. \citet{Chirici2013} compared the use of CART, RF, and Stochastic Gradient Boosting SGB, an ensemble tree method that uses both boosting and bagging, for mapping forest fuel types in Italy, and found that SGB had the highest overall accuracy.

\subsubsection{Fire detection}
Detecting wildfires as soon as possible after they have ignited, and therefore while they are still relatively small, is critical to facilitating a quick and effective response.  Traditionally, fires have mainly been detected by human observers, by distinguishing smoke in the field of view directly from a fire tower, or from a video feed from a tower, aircraft, or from the ground.  All of these methods can be limited by spatial or temporal coverage, human error, the presence of smoke from other fires and by hours of daylight. Automated detection of heat signatures or smoke in infra-red or optical images can extend the spatial and temporal coverage of detection, the detection efficiency in smoky conditions, and remove bias associated with human observation. The analytical task is a classification problem that is quite well suited to ML methods. 


For example, \citet{Arrue2000} used ANNs for infrared (IR) image processing (in combination with visual imagery, meteorological and geographic data used in a decision function using fuzzy logic), to identify true wildfires. Several researchers have similarly employed ANNs for fire detection \fixed{\citep{Al-Rawi2001, Angayarkkani2010,Fernandes2004,Fernandes2004a,Li2015,Soliman2010,Utkin2002,Sayad2019}}. In addition, \citet{Liu2015} used ANNs on wireless sensor networks to build a fire detection system, where multi-criteria detection was used on multiple attributes (e.g. flame, heat, light, and radiation) to detect and raise alarms. Other ML methods used in fire detection systems include SVM to automatically detect wildfires from videoframes \citep{Zhao2011}, GA for multi-objective optimization of a LiDAR-based fire detection system \citep{Cordoba2004}, BN in a vision-based early fire detection system \citep{Ko2010}, ANFIS \citep{ Angayarkkani2011,Wang2011}, and KM \citep{Srinivasa2008}.  

CNNs (ie. deep learning), which are able to extract features and patterns from spatial images and are finding widespread use in object detection tasks, have recently been applied to the problem of fire detection. Several of these applications trained the models on terrestrial based images of fire and/or smoke \citep{Zhang2016, Zhang2018, Zhang2018a, Yuan2018, Akhloufi2018, Barmpoutis2019, Jakubowski2019, JoaoSousa2019, Li2018a, Li2019, Muhammad2018, Wang2019}. Of particular note, \citet{Zhang2018a} found CNNs outperformed a SVM-based method and \citet{Barmpoutis2019} found a Faster region-based CNN outperformed another CNN based on YOLO \fixed{(``you only look once’’)}. \citet{Yuan2018} used CNN combined with optical flow to include time-dependent information. \citet{Li2018a} similarly used a 3D CNN to incorporate both spatial and temporal information and so were able to treat smoke detection as a segmentation problem for video images. Another approach by \citet{Cao2019} used convolutional layers as part of a Long Short Term Memory (LSTM) Neural network for smoke detection from a sequence of images (ie. video feed). They found the LSTM method achieved 97.8\% accuracy, a 4.4\% improvement over a single image-based deep learning method. 

Perhaps of greater utility for fire management were fire/smoke detection models trained on either unmanned aerial vehicle (UAV) images \citep{Zhao2018, Alexandrov2019} or satellite imagery including GOES-16 \citep{Phan2019} and MODIS \citep{Ba2019}.  \citet{Zhao2018} compared SVM, ANN and 3 CNN models and found their 15-layer CNN performed best with an accuracy of 98\%. By comparison, the SVM based method, which was unable to extract spatial features, only had an accuracy of 43\%. \citet{Alexandrov2019} found YOLO was both faster and more accurate than a region-based CNN method in contrast to \citet{ Barmpoutis2019}.

\subsubsection{Fire perimeter and severity mapping}

Fire maps have two management applications: 1) Accurate maps of the location of the active fire perimeter are important for daily planning of suppression activities and/or evacuations, including modeling fire growth 2) Maps of the final burn perimeter and fire severity are important for assessing and predicting the economic and ecological impacts of wildland fire and for recovery planning.  Historically, fire perimeters were sketch-mapped from the air, from a ground or aerial GPS or other traverse, or by air-photo interpretation.  Developing methods for mapping fire perimeters and burn severity from remote sensing imagery has been an area of active research since the advent of remote sensing in the 1970s, and is mainly concerned with classifying active fire areas from inactive or non burned areas, burned from unburned areas (for extinguished fires), or fire severity measures such as the Normalized Burn Ratio \citep{Lutes2006}. 


In early studies using ML methods for fire mapping \citet{Al-Rawi2001} and \citet{Al-Rawi2002b} used ANNs (specifically, the supervised ART-II neural network) for burned scar mapping and fire detection. \citet{Pu2004} compared Logistic Regression (LR) with ANN for burned scar mapping using Landsat images; both methods achieved high accuracy ($>97\%$). Interestingly, however, the authors found that LR was more efficient for their relatively limited data set. The authors in \citet{Zammit2006} performed burned area mapping for two large fires that occurred in France using satellite images and three ML algorithms, including SVM, K-nearest neighbour, and the K-means algorithm; overall SVM had the best performance. Likewise, \citet{Dragozi2011} compared the use of SVM against a nearest neighbour method for burned area mapping in Greece and found better performance with SVM. In fact, a number of studies \citep{Alonso-Benito2008,Cao2009,Petropoulos2010,Petropoulos2011,Zhao2015,Pereira2017,Branham2017,Hamilton2017} have successfully used SVM for burned scar mapping using satellite data. \citet{Mitrakis2012} performed burned area mapping in the Mediterranean region using a variety of ML algorithms, including a fuzzy neuron classifier (FNC), ANN, SVM, and AdaBoost, and found that, while all methods displayed similar accuracy, the FNC performed slightly better. \citet{Dragozi2014} applied SVM and a feature selection method (based on fuzzy logic) to IKONOS imagery for burned area mapping in Greece. Another approach to burned area mapping in the Mediterranean used an ANN and MODIS hotspot data \citep{Gomez2011}. \citet{Pereira2017} used a one class SVM, which requires only positive training data (i.e. burned pixels), for burned scar mapping, which may offer a more sample efficient approach than general SVMs -- the one class SVM approach may be useful in cases where good wildfire training datasets are difficult to obtain. In \citet{Mithal2018}, the authors developed a three-stage framework for burned area mapping using MODIS data and ANNs. \citet{Crowley2019} used Bayesian Updating of Landcover (BULC) to merge burned-area classifications from three remote sensing sources (Landsat-8, Sentinel-2 and MODIS). \citet{Celik2010} used GA for change detection in satellite images, while \citet{Sunar2001} used the interactive Iterative Self-Organizing DATA algorithm (ISODATA) and ANN to map burned areas. 

	In addition to burned area mapping, ML methods have been used for burn severity mapping, including GA \citep{Brumby2001}, MaxEnt \citep{Quintano2019}, bagged decision trees \citep{Sa2003}, and others. For instance, \citet{Hultquist2014a} used three popular ML approaches (Gaussian Process Regression (GPR) \citep{rasmussen2006gaussian}, RF, and SVM) for burn severity assessment in the Big Sur ecoregion, California. RF gave the best overall performance and had lower sensitivity to different combinations of variables. All ML methods, however, performed better than conventional multiple regression techniques. Likewise, \citet{Hultquist2014a} compared the use of GPR, RF, and SVM for burn severity assessment, and found that RF displayed the best performance. Another recent paper by \citet{Collins2018} investigated the applicability of RF for fire severity mapping, and discussed the advantages and limitations of RF for different fire and land conditions.
	
One recent paper by \citet{Langford2019} used a 5-layer deep neural network (DNN) for mapping fires in Interior Alaska with a number of MODIS derived variables (eg. NDVI and surface reflectance). They found that a validation-loss (VL) weight selection strategy for the unbalanced data set (i.e., the no-fire class appeared much more frequently than fire) allowed them to achieve better accuracy compared with a XGBoost method. However, without the VL approach, XGBoost outperformed the DNN, highlighting the need for methods to deal with unbalanced datasets in fire mapping.

\subsection{Fire Weather and Climate Change} 

\subsubsection{Fire weather prediction}

Fire weather is a critical factor in determining whether a fire will start, how fast it will spread, and where it will spread. Fire weather observations are commonly obtained from surface weather station networks operated by meteorological services or fire management agencies. Weather observations may be interpolated from these point locations to a grid over the domain of interest, which may include diverse topographical conditions; the interpolation task is a regression problem. Weather observations may subsequently be used in the calculation of meteorologically based fire danger \fixed{indices}, such as the Canadian Fire Weather Index (FWI) System \citep{VanWagner1987}.  Future fire weather conditions and danger \fixed{indices} are commonly forecast using the output from numerical weather prediction (NWP) models (e.g., The European Forest Fire Information System \citep{ San-Miguel-Ayanz2012}). However, errors in the calculation of fire danger \fixed{indices} that have a memory (such as the moisture \fixed{indices} of the FWI System) can accumulate in such projections. It is noteworthy that surface fire danger measures may be correlated with large scale weather and climatic patterns.  

To date there has been relatively few papers that address fire weather and danger prediction using machine learning. The first effort \citep{Crimmins2006} used self-organizing maps (SOMs) to explore the synoptic climatology of extreme fire weather in the southwest USA. He found three key patterns representing southwesterly flow and large geopotential height gradients that were associated with over 80\% of the extreme fire weather days as determined by a fire weather index. \citet{Nauslar2019} used SOMs to determine the timing of the North American Monsoon that plays a major role on the length of the active fire season in the southwest USA. \citet{Lagerquist2017} also used SOMs to predict extreme fire weather in northern Alberta, Canada. Extreme fire weather was defined by using extreme values of the Fine Fuel Moisture Code (FFMC), Initial Spread Index (ISI) and the Fire Weather Index (FWI), all components of the Canadian Fire Weather Index (FWI) System \citep{VanWagner1987}. Good performance was achieved with the FFMC and the ISI and this approach has the potential to be used in near real time, allowing input into fire management decision systems. Other efforts have used a combination of conventional and machine learning approaches to interpolate meteorological fire danger in Australia \citep{Sanabria2013}. 

\subsubsection{Lightning prediction}
\fixed{Lightning is second most common cause of wildfires (behind human causes);} thus predicting the location and timing of future storms/strikes is of great importance to predicting fire occurrence. Electronic lightning detection systems have been deployed in many parts of the world for several decades and have accrued rich strike location/time datasets. Lightning prediction models have employed these data to derive regression relationships with atmospheric conditions and stability indices that can be forecast with NWP. 
Ensemble forecasts of lightning using RF is a viable modelling approach for Alberta, Canada \citep{Blouin2016}. \citet{Bates2017} used two machine learning methods (CART and RF) and three statistical methods to classify wet and dry thunderstorms (lightning associated with dry thunderstorms are more likely to start fires) in Australia.

\subsubsection{Climate Change}
\index{Climate Change}
Transfer modeling, whereby a model produced for one study region and/or distribution of environmental conditions is applied to other cases \citep{Phillips2006}, is a common approach in climate change science.  Model transferability should be considered when using ML methods to estimated projected quantities due to climate change or other environmental changes. With regards to climate change, transfer modeling is essentially an extrapolation task. Previous studies in the context of species distribution modeling have indicated ML approaches may be suitable for transfer modeling under future climate scenarios. For example, \citet{Heikkinen2012} indicated MaxEnt and generalized boosting methods (GBM) have the better transferability than either ANN and RF, and that the relatively poor transferability of RF may be due to overfitting.  

There are several publications on wildfires and climate change that use ML approaches. 
\citet{Amatulli2013} found that Multivariate Adaptive Regression Splines (MARS) were better predictors of future monthly area burned for 5 European countries as compared to Multiple Linear Regression and RF. \citep{Parks2016} projected fire severity for future time periods in Western USA using BRT. \citet{Young2017} similarly used BRT to project future fire intervals in Alaska and found up to a fourfold increase in (30 year) fire occurrence probability by 2100. Several authors used MaxEnt to project future fire probability globally \citep{Moritz2012}, for Mediterranean ecosystems \citep{Batllori2013},  in Southwest China \citep{Li2017}, the pacific northwestern USA \citep{Davis2017}, and for south central USA \citep{Stroh2018}. 
An alternative approach for projecting future potential burn probability was employed by \citet{Stralberg2018a} who used RF to determine future vegetation distributions as inputs to ensemble Burn-P3 simulations. 
Another interesting paper of note was by \citet{Boulanger2018} who built a consensus model with 2 different predictor datasets and 5 different regression methods (generalised linear models, RF, BRT, CART and MARS) to make projections of future area burned in Canada. The consensus model can be used to quantify uncertainty in future area burned estimates. The authors noted that model uncertainty for future periods ($> 200\%$) can be higher than that of different climate models under different carbon forcing scenarios. This highlights the need for further work in the application of ML methods for projecting future fire danger under climate change. 

\subsection{Fire Occurrence, Susceptibility and Risk}

Papers in this domain include prediction of fire occurrence and area burned (at a landscape or seasonal scales), mapping of fire susceptibility (or similar definitions of risk) and analysis of landscape or environmental controls on fire.

\subsubsection{Fire occurrence prediction}

Predictions of the number and location of fire starts in the upcoming day(s) are important to preparedness planning --- that is, the acquisition of resources, including the relocation of mobile resources and readiness for expected fire activity.  The origins of fire occurrence prediction (FOP) models go back almost 100 years \citep{Nadeem2020}. FOP models typically use regression methods to relate the response variable (fire reports or hotspots) to weather, lightning, and other covariates for a geographic unit, or as a spatial probability. The seminal work of Brillinger and others in developing the spatio-temporal FOP framework is reviewed in \citet{Taylor2013} 
The most commonly used ML method in studies predicting fire occurrence were ANNs. As early as 1996, \citet{Vega-Garcia1996} used an ANN for human-caused wildfire prediction in Alberta, Canada, correctly predicting 85\% of no-fire observations and 78\% of fire observations. Not long after, \citet{Alonso-Betanzos2002} and \citet{Alonso-Betanzos2003} used ANN to predict a daily fire occurrence risk index using temperature, humidity, rainfall, and fire history, as part of a larger system for real-time wildfire management system in the Galicia region of Spain. \citet{Vasilakos2007} used separate ANNs for three different indices representing fire weather (Fire Weather Index; FWI), hazard (Fire Hazard Index; FHI), and risk (Fire Risk Index) to create a composite fire ignition index (FII) for estimating the probability of wildfire occurrence on the Greek island of Lesvos. \citet{Sakr2010} used meteorological variables in a SVM to create a daily fire risk index corresponding to the number of fires that could potentially occur on a particular day. \citet{Sakr2011} then compared the use of SVM and ANN for fire occurrence prediction based only on relative humidity and cumulative precipitation up to the specific day. While \citet{Sakr2011} reported low errors for the number of fires predicted by both the SVM and ANN models, ANN models outperformed SVM; however, the SVM performed better on binary classification of fire/no fire. It is important to note, however, that ANNs encompass a wide range of possible network architectures. In an Australian study, \citet{Dutta2013} compared the use of ten different types of ANN models for estimating monthly fire occurrence from climate data, and found that an Elman RNN performed the best.

	After 2012, RF became the more popular method for predicting fire occurrence among the papers reviewed here. \citet{Stojanova2012} evaluated several machine learning methods for predicting fire outbreaks using geographical, remote sensed, and meteorological data in Slovenia, including single classifier methods (i.e., KNN, Naive Bayes, DT (using the J48 and jRIP algorithms), LR, SVM, and BN), and ensemble methods (AdaBoost, DT with bagging, and RF). The ensemble methods DT with bagging and RF displayed the best predictive performance with bagging having higher precision and RF having better recall. \citet{Vecin-Arias2016} found that RF performed slightly better than LR for predicting lightning fire occurrence in the Iberian Peninsula, based on topography, vegetation, meteorology, and lightning characteristics. Similarly, \citet{Cao2017} found that a cost-sensitive RF analysis outperformed GLM and ANN models for predicting wildfire ignition susceptibility. In recent non-comparative studies, \citet{Yu2017} used RF to predict fire risk ratings in Cambodia using publicly available remote sensed products, while \citet{VanBeusekom2018} used RF to predict fire occurrence in Puerto Rico and found precipitation was found to be the most important predictor.
	The maximum entropy (MaxEnt) method has also been used for fire occurrence prediction \citep{DeAngelis2015,Chen2015}. For example, \citet{DeAngelis2015} used MaxEnt to evaluate  different meteorological variables and fire-indices (e.g. the Canadian Fire Weather Index, FWI) for daily fire risk forecasting in the mountainous Canton Ticino region of Switzerland. The authors of that study found that combinations of such variables increased predictive power for identifying daily meteorological conditions for wildfires.  
\citet{Dutta2016} use a two-stage machine learning approach (ensemble of unsupervised deep belief neural networks with conventional supervised ensemble machine learning) to predict bush-fire hot spot incidence on a weekly time-scale. In the first unsupervised deep learning phase, \citet{Dutta2016} used Deep Belief Networks (DBNet; an ensemble deep learning method) to generate simple features from environmental and climatic surfaces. In the second supervised ensemble classification stage, features extracted from the first stage were fed as training inputs to ten ML classifiers (i.e., conventional supervised Binary Tree, Linear Discriminant Analyser, Naïve Bayes, KNN, Bagging Tree, AdaBoost, Gentle Boosting Tree, Random Under-Sampling Boosting Tree, Subspace Discriminant, and Subspace KNN) to establish the best classifier for bush fire hotspot estimation. The authors found that bagging and the conventional KNN classifier were the two best classifiers with 94.5\% and 91.8\% accuracy, respectively.

\subsubsection{Landscape scale burned area prediction}

The use of ML methods in studies of burned area prediction have only occurred relatively recently compared to other wildfire domains, yet such studies have incorporated a variety of ML methods. For example, \citet{Cheng2008} used an RNN to forecast annual average area burned in Canada, while \citet{Archibald2009} used RF to evaluate the relative importance of human and climatic drivers of burnt area in Southern Africa. \citet{Arnold2014} used Hard Competitive Learning (HCL) to identify clusters of unique pre-fire antecedent climate conditions in the interior western US which they then used to construct fire danger models based on MaxEnt.

	\citet{Mayr2018} evaluated five common statistical and ML methods for predicting burned area and fire occurrence in Namibia, including GLM, Multivariate Adaptive Regression Splines (MARS), Regression Trees from Recursive Partitioning (RPART), RF, and SVMs for Regression (SVR). The RF model performed best for predicting burned area and fire occurrence; however, adjusted R$^2$ values were slightly higher for RPART and SVR in both cases. Likewise, \citet{deBem2019} compared the use of LR and ANN for modelling burned area in Brazil. Both LR and ANN showed similar performance; however, the ANN had better accuracy values when identifying non-burned areas, but displayed lower accuracy when classifying burned areas.

\subsubsection{Fire Susceptibility Mapping}

A considerable number of references (71) used various ML algorithms to map wildfire susceptibility, corresponding to either the spatial probability or density of fire occurrence (or other measures of fire risk such as burn severity) although other terms such as fire vulnerability and risk have also been used. The general approach was to build a spatial fire susceptibility model using either remote sensed or agency reported fire data with some combination of landscape, climate, structural and anthropogenic variables as explanatory variables. In general, the various modeling approaches used either a presence only framework (e.g., MaxEnt) or a presence/absence framework (e.g., BRT or RF).

Early attempts at fire susceptibility mapping used CART \citep{Amatulli2006,  Amatulli2007, Lozano2008}. \citet{Amatulli2007} compared fire density maps in central Italy using CART and multivariate adaptive regression splines (MARS) and found while CART was more accurate that MARS led to smoother density model. More recent work has used ensemble based classifiers, such as RF and BRT, or ANNs (see \fixed{table S.3.3 in supplementary material for a full list})   Several of these papers also compared ML and non-ML methods for fire susceptibility mapping and in general found superior performance from the ML methods. Specifically, \citet{Adab2017} mapped fire hazard in the Northeast of Iran, and found ANN performed better than binary logistic regression (BLR) with an AUC of 87\% compared with 81\% for BLR. \citet{Bisquert2012} found ANN outperformed logistic regression for mapping fire risk in the North-west of Spain. \citet{Goldarag2016} also compared ANN and linear regression for fire susceptibility mapping in Northern Iran and found ANN had much better accuracy (93.49\%) than linear regression (65.76\%). \citet{Guo2016} and \citet{Guo2016a} compared RF and logistic regression for fire susceptibility mapping in China and found RF led to better performance. \citet{Oliveira2012} compared RF and LR for fire density mapping in Mediterranean Europe and found RF outperformed linear regression.  \citet{PerestrelloDeVasconcelos2001} found ANN had better classification accuracy than logistic regression for ignition probability maps in parts of Portugal. 

Referring to \fixed{table \ref{tab:summary} and section S.3.3 of the supplementary material} a frequently used ML method for fire susceptibility mapping was Maximum Entropy (MaxEnt) which is extensively used in landscape ecology for species distribution modeling \citep{Elith2011}. 
 In particular, \citet{Vilar2016} found MaxEnt performed better than GLM for fire susceptibility mapping in central Spain with respect to sensitivity (i.e., true positive rate) and commission error (i.e., false positive rate), even though the AUC was lower. Of further note, \citet{Duane2015} partitioned their fire data into topography-driven, wind-driven and convection-driven fires in Catalonia and mapped the fire susceptibility for each fire type. 

Other ML methods used for regional fire susceptibility mapping include Bayesian networks \citep{Bashari2016, Dlamini2011} and novel hybrid methods such as Neuro-Fuzzy systems \citep{Jaafari2019a, TienBui2017}. \citet{Bashari2016} noted that Bayesian networks may be useful because it allows probabilities to be updated when new observations become available. SVM was also used by a number of authors as a benchmark for other ML methods \citep{Ghorbanzadeh2019b, Gigovic2019, Hong2018, Jaafari2019, NgocThach2018, Rodrigues2014a, Sachdeva2018, Tehrany2018a, TienBui2017, VanBreugel2016, Zhang2019} but as we discuss below, it did not perform as well as other methods to which it was being compared. 

There were two applications of ML for mapping global fire susceptibility including \citet{Moritz2012} who used MaxEnt and \citet{Luo2013} who used RF. Both of these papers found that at a global scale, precipitation was one of the most important predictors of fire risk.

The majority of papers considered thus far used the entire study period (typically 4 or more years) to map fire susceptibility, therefore neglecting the temporal aspect of fire risk. However, a few authors have considered various temporal factors to map fire susceptibility. \citet{Martin2019} included seasonality and holidays as explanatory variables for fire probability in northeast Spain. \citet{Vacchiano2018} predicted fire susceptibility separately for the winter and summer seasons. Several papers produced maps of fire susceptibility in the Eastern US by month of year \citep{Peters2013, Peters2017}. \citet{Parisien2014} examined differences in annual fire susceptibility maps and a 31 year climatology for the USA, highlighting the role of climate variability as a driver of fire occurrence. In particular, they found FWI90 (the 90th percentile of the Canadian Fire Weather Index) was the dominant factor for annual fire risk but not for climatological fire risk. \citet{Cao2017} considered a 10 day resolution (corresponding to the available fire data) for fire risk mapping, which makes their approach similar to fire occurrence prediction.  

In addition to fire susceptibility mapping, a few papers focused on other aspects of fire risk including mapping probability of burn severity classes \citep{Holden2009, Parks2018, Tracy2018}. \citet{Parks2018} additionally considered the role of fuel treatments on fire probability which has obvious implications for fire management. Additionally \citet{Ghorbanzadeh2019a} combined fire susceptibility maps with vulnerability and infrastructure indicators to produce a fire hazard map. 

A number of papers directly compared three or more ML (and sometimes non-ML) methods for fire susceptibility mapping. Here we highlight some of these papers, which elucidate the performance and advantages/disadvantages of various ML methods.  \citet{Cao2017} found a cost-sensitive RF model outperformed a standard RF model, ANN as well as probit and logistic regression. \citet{Ghorbanzadeh2019b} compared ANN, SVM and RF and found the best performance with RF. \citet{Gigovic2019} compared SVM and RF for fire susceptibility mapping in combination with Bayesian averaging to generate ensemble models. They found the ensemble model led to marginal improvement (AUC = 0.848) over SVM (AUC=0.834) and RF (AUC=0.844). 
For mapping both wildfire ignitions and potential natural vegetation in Ethiopia \citet{VanBreugel2016} also considered ensemble models consisting of a weighted combination of ML methods (RF, SVM, BRT, MaxEnt, ANN, CART) and non-ML methods (GLM and MARS) and concluded the ensemble member performed best over a number of metrics. However, in this paper RF showed the best overall performance of all methods including the ensemble model. 

\citet{Jaafari2018} compared 5 decision tree based classifiers for wildfire susceptibility mapping in Iran. Here, the Alternating Decision tree (ADT) classifier achieved the highest performance (accuracy 94.3\%) in both training and validation sets. \citet{NgocThach2018} compared SVM, RF and a Multilayer Perceptron (MLP) neural network for forest fire danger mapping in the region of Tjuan chau in Vietnam. They found the performance of all models were comparable although MLP had the highest AUC values.
Interestingly \citet{Pourtaghi2016} found that a generalized additive model (GAM) outperformed RF and BRT for fire susceptibility mapping in the Golestan province in Iran. This was one of the few examples we found where a non-ML method outperformed ML methods.
\citet{Rodrigues2014a} compared RF, BRT, SVM and logistic regression for fire susceptibility mapping and found RF led to the highest accuracy as well as the most parsimonious model. 
\citet{Tehrany2018a} compared a LogitBoost ensemble-based decision tree (LEDT) algorithm with SVM, RF and Kernel logistic regression (KLR) for fire susceptibility mapping in Lao Cai region of Vietnam and found the best performance with LEDT, closely followed by RF.  Finally, of particular note, \citet{Zhang2019} compared CNN, RF, SVM, ANN and KLR for fire susceptibility mapping in the Yunnan Province of China. This was the only application of deep learning we could find for fire susceptibility mapping. The authors found that CNN outperformed the other algorithms with overall accuracy of 87.92\% compared with RF (84.36\%), SVM (80.04\%), MLP (78.47\%), KLR (81.23\%). They noted that the benefit of CNN is that it incorporates spatial correlations so that it can learn spatial features. However, the downside is that deep learning models are not as easily interpretted as other ML methods (such as RF and BRT).

 \subsubsection{Landscape controls on fire}\label{sec:controls}
 
 Many of the ML methods used in fire susceptibility mapping have also been used to examine landscape controls -- ie. the relative importance of weather, vegetation, topography, structural and anthropogenic variables -- on fire activity, which may facilitate hypothesis formation and testing or model building. From table \ref{tab:summary} the most commonly used methods in this section were MaxEnt, RF, BRT and ANN. These methods all allow for the determination of variable importance (i.e. the relative influence of predictor variables in a given model of a response variable). A commonly used method to ascertain variable importance is through the use of partial dependence plots \citep{Hastie2009}. This method works by averaging over models that exclude the predictor variable of interest, with the resulting reduction in AUC (or other performance metrics) representing the marginal effect of the variable on the response. Partial dependence plots have the advantage of being able to be applied to a wide range of ML methods. A related method for determining variable importance, often used for RFs, is a permutation test which involves random permutation of each predictor variable \citep{Strobl2007}. Another model-dependent approach used for ANN is the use of partial derivatives (of the activation functions of hidden and output nodes) as outlined by \citet{Vasilakos2009}. It should be noted that while many other methods for model interpretation and variable dependence exist, a discussion of these methods is outside the scope of this paper. 

In general, the drivers of fire occurrence or area burned varied greatly by the study area considered (including the size of area) and the methods used. Consistent with other work on ``top down'' and ``bottom up'' drivers of fire activity, at large scales climate variables were often determined to be the main drivers of fire activity whereas at smaller scales anthropogenic or structural factors exerted a larger influence. Here we discuss some of the papers that highlight the diversity of results for different study areas and spatial scales (global, country, ecoregion, urban) but refer the reader to \fixed{section S.3.4 of the supplementary material} for a full listing of papers in this section. Note that many of the papers listed under \fixed{section S.3.4}  also belong to the fire susceptibility mapping section and have already been discussed there. 

\citet{Aldersley2011} considered drivers of monthly area burned at global and regional scales using both regression trees and RF. They found climate factors (high temperature, moderate precipitation, and dry spells) were the most important drivers at the global scale, although at the regional scale the models exhibited higher variability due to the influence of anthropogenic factors. At a continental scale \citet{Mansuy2019} used MaxEnt to show that climate variables were the dominant controls (over landscape and human factors) on area burned for most ecoregions for both protected areas and outside these areas, although anthropogenic factors exerted a stronger influence in some regions such as the Tropical Wet Forests ecoregion. \citep{Masrur2018} used RF to investigate controls on circumpolar arctic fire and found June surface temperature anomalies were the most important variable for determining the likelihood of wildfire occurrence on an annual scale. \citet{Chingono2015} used MaxEnt to model fire occurrences in Southern Africa where most fires are human-caused and found vegetation (i.e., dry mass productivity and NDVI) were the main drivers of biomass burning. 
\citet{Curt2015} used BRT to examine drivers of fire in New Caledonia. Interestingly, they found that human factors (such as distance to villages, cities or roads) were dominant influences for predicting fire ignitions whereas vegetation and weather factors were most important for area burned. \citet{Curt2016} modeled fire probabilities by different fire ignition causes (lightning, intentional, accidental, negligence professional and negligence personal) in Southeastern France. They found socioeconomic factors (eg. housing and road density) were the dominant factors for ignitions and area burned for human-caused fires. \citet{Fernandes2016} used BRT to examine large fires in Portugal and found high pyrodiversity (ie.  spatial structure due to fire recurrence) and low landscape fuel connectivity were important drivers of area burned. \citet{Curt2016} modeled fire probabilities by different fire ignition causes (lightning, intentional, accidental, negligence professional and negligence personal) in Southeastern France. They found socioeconomic factors (eg. housing and road density) were the dominant factors for ignitions and area burned for human-caused fires. \citet{Leys2017} used RF to find the drivers that determine sedimentary charcoal counts in order to reconstruct grassfire history in the Great Plains, USA. Not surprisingly, they found fire regime characteristics (eg. area burned and fire frequency) were the most important variables and concluded that charcoal records can therefore be used to reconstruct fire histories. \citet{Li2009} used ANNs to show that wildfire probability was strongly influenced by population density in Japan, with a peak determined by the interplay of positive and negative effects of human presence. This relationship, however, becomes more complex when weather parameters and forest cover percentage are added to the model. \citet{Liu2013a} used BRT to study factors influencing fire size in the Great Xingan Mountains in Northeastern China. Their method included a ``moving window'' resampling technique that allowed them to look at the relative influence of variables at different spatial scales. They showed that the most dominant factors influencing fire size were fuel and topography for small fires, but fire weather became the dominant factor for larger fires. For regions of high population density, anthropogenic or structural factors are often dominant for fire susceptibility. For example \citet{Molina2019} used MaxEnt to show distance to roads, settlements or powerlines were the dominant factors for fire occurrence probability in the Andalusia region in southern Spain. 
MaxEnt has also been used for estimating spatial fire probability under different scenarios such as future projections of housing development and private land conservation \citep{Syphard2016}. One study in China using RF found mean spring temperature was the most important variable for fire occurrence whereas forest stock was most important for area burned \citep{Ying2018}. 

Some authors examined controls on fire severity using high resolution data for a single large fire. For example, several authors used RF to examine controls on burn severity for the 2013 Rim fire in the Sierra Nevada \citep{Lydersen2014, Kane2015, Lydersen2017}. At smaller spatial scales fire weather was the most important variable for fire severity, whereas fuel treatments were most important at larger spatial scales \citep{Lydersen2017}. A similar study by \citet{Harris2017} showed that previous fire severity was an important factor influencing fire severity for the Rim fire. For the 2005 Riba de Saelices fire, \citet{Viedma2015} looked at factors contributing to burn severity using a BRT model and found burning conditions (including fire weather variables) were more important compared than stand structure and topography.  For burn severity these papers all used the Relativized differenced Normalized Burn Ratio (RdNBR) metric, derived from Landsat satellite images, which allowed spatial modeling at high resolutions (eg. 30m by 30m).  
In addition to the more commonly used ML methods one paper by \citet{Wu2015} used KNN to identify spatially homogeneous fire environment zones by clustering climate, vegetation, topography, and human activity related variables. They then used CART to examine variable importance for each of three fire environment zones in south-eastern China. 
For landscape controls on fire there were few studies comparing multiple ML methods. One such study by \citet{Nelson2017} compared CART, BRT and RF for classifying different fire size classes in British Columbia, Canada. For both central and periphery regions they found the best performing model was BRT followed by CART and RF. For example, in the central region BRT achieved a classification accuracy of 88\% compared with 82.9\% and 49.6\% for the CART and RF models respectively. It is not clear from the study why RF performed poorly, although it was noted that variable importance differs appreciably between the three models.

\subsection{Fire Behavior Prediction}

 In general, fire behavior includes physical processes and characteristics at a variety of scales including combustion rate, flaming, smouldering residence time fuel consumption, flame height, and flame depth. However, the papers in this section deal mainly with larger scale processes and characteristics such as the prediction of fire spread rates, fire growth, burned area, and fire severity, conditional on the occurrence (ignition) of one, or more, wildfires. Here, our emphasis is on prognostic applications, in contrast to the \textit{Fuels Characterization, Fire Detection and Mapping} problem domain, in which we focused on diagnostic applications.

\subsubsection{Fire spread and growth}

Predicting the spread of a wildland fire is an important task for fire management agencies, particularly to aid in the deployment of suppression resources or to anticipate evacuations one or more days in advance. Thus, a large number of models have been developed using different approaches.  In a series of reviews \citet{Sullivan2009,Sullivan2009a,Sullivan2009b} described fire spread models he classified as being of physical or quasi-physical nature, or empirical or quasi-empirical nature, as well as mathematical analogues and simulation models. Many fire growth simulation models convert one dimensional empirical or quasi-empirical spread rate models to two dimensions and then propagate a fire perimeter across a modelled landscape. 
 
 A wide range of ML methods have been applied to predict fire growth. For example, \citet{Markuzon2009} tested several classifiers (RF, BNs, and KNN) to estimate if a fire would become large either one or two days following its observation; they found each of the tested methods performed similarly with RF correctly classifying large fires at a rate over 75\%, albeit with a number of false positives.  \citet{Vakalis2004} used a ANN in combination with a fuzzy logic model to estimate the rate of spread in the mountainous region of Attica in Greece.
A number of papers used genetic algorithms (GAs) to optimize input parameters to a physics or empirically based fire simulator in order to improve fire spread predictions \citep{Abdalhaq2005, Rodriguez2008, Rodriguez2009, Artes2014, Artes2016, Carrillo2016, Denham2012, Cencerrado2012, Cencerrado2013, Cencerrado2014, Artes2017, Denham2018}. For example, \citet{Cencerrado2014} developed a framework based on GAs to shorten the time needed to run deterministic fire spread simulations. They tested the framework using the FARSITE \citep{Finney2004} fire spread simulator with different input scenarios sampled from distributions of vegetation models, wind speed/direction, and dead/live fuel moisture content. The algorithm used a fitness function which discarded the most time-intensive simulations, but did not lead to an appreciable decrease in the accuracy of the simulations. Such an approach is potentially useful for fire management where it is desirable to predict fire behavior as far in advance as possible so that the information can be enacted upon. This approach may greatly reduce overall simulation time by reducing the input parameter space as also noted by \citet{Artes2016} and \citet{Denham2012}, or through parallelization of simulation runs for stochastic approaches \citep{Artes2017,Denham2018}. A different goal was considered by \citet{Ascoli2015} who used a GA to optimize fuel models in Southern Europe by calibrating the model with respect to rate of spread observations.  

\citet{Kozik2013} presented a fire spread model that used a novel ANN implementation that incorporated a Kalman filter for data assimilation that could potentially be run in real-time, the resulting model more closely resembling that of complex cellular automata than a traditional ANN. The same authors later implemented this model and simulated fire growth under various scenarios with different wind speeds and directions, or both, although a direct comparison with real fire data was not possible \citep{Kozik2014}. 

\citet{Zheng2017} simulated fire spread by integrating a cellular automata (CA) model with an Extreme Learning Machine (ELM; a type of feedforward ANN). Transition rules for the CA were determined by the ELM trained with data from historical fires, as well as vegetation, topographic, and meteorological data. Likewise, \citet{Chetehouna2015} used ANNs to predict fire behavior, including rate of spread, and flame height and angle. In contrast, \citet{Subramanian2017} formulated the problem of fire spread prediction as a Markov Decision Process, where they proposed solutions based on both a classic reinforcement learning algorithm and a deep reinforcement learning algorithm -- the authors found the deep learning approach improved on the traditional approach when tested on two large fires in Alberta, Canada. The authors further developed this work to compare five widely used reinforcement learning algorithms \citep{GanapathiSubramanian2018}, and found that the Asynchronous Advantage Actor-Critic (A3C) and Monte Carlo Tree Search (MCTS) algorithms achieved the best accuracy. Meanwhile, \citet{Khakzad2019} developed a fire spread model to predict the risk of fire spread in Wildland-Industrial Interfaces, using Dynamic Bayesian Networks (DBN) in combination with a deterministic fire spread model. The Canadian Fire Behavior Prediction (FBP) system, which uses meteorological and fuel conditions data as inputs, determined the fire spread probabilities from one node to another in the aforementioned DBN. 

More recently \citet{Hodges2019} trained a (deep learning) CNN to predict fire spread using environmental variables (topography, weather and fuel related variables). Outputs of the CNN were spatial grids corresponding to the probability the burn map reached a pixel and the probability the burn map did not reach a pixel. Their method achieved a mean precision of 89\% and mean sensitivity of 80\% with reference 6 hourly burn maps computed using the physics-based FARSITE simulator. \citet{Radke2019} also used a similar approach to predict daily fire spread for the 2016 Beaver Creek fire in Colorado.

\subsubsection{Burned area and fire severity prediction}

There are a number of papers that focus on using ML approaches to directly predict the final area burned from a wildfire. \citet{Cortez2007} compared multiple regression and four different ML methods (DT, RF, ANN, and SVM) to predict area burned using fire and weather (i.e., temperature, precipitation, relative humidity and wind speed) data from the Montesinho natural park in northeastern Portugal, and found that SVM displayed the best performance. A number of publications subsequently used the data from \citet{Cortez2007} to predict area burned using various ML methods, including ANN \citep{Safi2013, Storer2016}, genetic algorithms \citep{Castelli2015a}, both ANN and SVM \citep{AlJanabi2018}, and decision trees \citep{Alberg2015, Li2018}. Notably, \citet{Castelli2015a} found that a GA variant outperformed other ML methods including SVM. \citet{Xie2014} used a similar set of input variables with SVM to predict burned area in for Guangzhou City in China. In addition to these studies, \citet{Toujani2018} used hidden Markov models (HMM) to predict burned area in the north-west of Tunisia, where the spatiotemporal factors used as inputs to the model were initially clustered using self-organizing maps (SOMs). \citet{Liang2019} compared back-propagation neural networks, recurrent neural networks (RNN) and Long Short Term Memory (LSTM) neural networks to predict wildfire scale, a quantity related to area burned and fire duration, in Alberta Canada. They found the highest accuracy (90.9\%) was achieved with LSTM. 

Most recently, \citet{Xie2019} compared a number of machine learning methods for estimating area burned (regression) and binary classification of fire sizes ($>$ 5 Ha) in Montesinho natural park, Portugal. For the regression task, they found a tuned RF algorithm performed better than standard RF, tuned and standard gradient boosted machines, tuned and standard generalized linear models (GLMs) and deep learning. For the classification problem they found extreme gradient boosting and deep learning had a higher accuracy than CART, RF, SVM, ANN, and logistic regression.  

By attempting to predict membership of burned area size classes, a number of papers were able to recast the problem of burned area prediction as a classification problem. For example, \citet{Yu2011} used a combination of SOMs and back-propagation ANNs to classify forest fires into size categories based on meteorological variables. This approach gave \citet{Yu2011} better accurary (~90\%) when compared with a rules-based method (~82\%). \citet{Ozbayoglu2012} estimated burned area size classes using geographical and meteorological data using three different machine learning methods: i) Multilayer Perceptron (MLP); ii) Radial Basis Function Networks (RBFN); and iii) SVM. Overall, the best performing method was MLP, which achieved a 65\% success rate, using humidity and windspeed as predictors. \citet{Zwirglmaier2013} used a BN to predict area burned classes using historical fire data, fire weather data, fire behaviour indices, land cover, and topographic data. \citet{Shidik2014} used a hybrid model (Fuzzy C-Means and Back-Propagation ANN) to estimate fire size classes using data from \citet{Cortez2007}, where the hybrid model performed best with an accuracy of 97.50\% when compared with Naive Bayes (55.5\%), DT (86.5\%), RF (73.1\%), KNN (85.5\%) and SVM (90.3\%). \citet{Mitsopoulos2017} compared BRT, RF and Logistic Regression to predict 3 burned area classes for fires in Greece. They found RF led to the best performance of the three tested methods and that fire suppression and weather were the two most important explanatory variables. \citet{Coffield2019} compared CART, RF, ANN, KNN and gradient boosting to predict 3 burned area classes at time of ignition in Alaska. They found a parsimonious model using CART with Vapor Pressure Deficit (VPD) provided the best performance of the models and variables considered. 

We found only one study that used ML to predict fire behavior related to fire severity, which is important in the context of fire ecology, suggesting that there are opportunities to apply ML in this domain of wildfire science. In that paper, \citet{Zald2018} used RF to determine that the most important predictor of fire severity was daily fire weather, followed by stand age and ownership, with less predictability given by topographic features.

\subsection{Fire Effects}

Fire Effects prediction studies have largely used regression based approaches to relate costs, losses, or other impacts (e.g., soils, post-fire ecology, wildlife, socioeconomic factors) to physical measures of fire severity and exposure. Importantly, this category also includes wildfire smoke and particulate modelling (but not smoke detection which was previously discussed in the fire detection section). 

\subsubsection{Soil Erosion and Deposits}
\citet{Mallinis2009} modelled potential post-fire soil erosion risk following a large intensive wildfire in the Mediterranean area using CART and k-means algorithms.  In that paper, before wildfire, 55\% of the study area was classified as having severe or heavy erosion potential, compared to 90\% post-fire, with an overall classification accuracy of 86\%. Meanwhile, \citet{Buckland2019} used ANNs to examine the relationships between sand deposition in semi-arid grasslands and wildfire occurrence, land use, and climatic conditions. The authors then predicted soil erosion levels in the future given climate change assumptions.

\subsubsection{Smoke and Particulate Levels}
Smoke emitted from wildfires can seriously lower air quality with adverse effects on the health of both human and non-human animals, as well as other impacts. Thus, it is not surprising that ML methods have been used to understand the dynamics of smoke from wildland fire. For example, \citet{Yao2018} used RF to predict the minimum height of forest fire smoke using data from the CALIPSO satellite. More commonly, ML methods have also been used to estimate population exposure to fine particulate matter (e.g., PM2.5: atmospheric particulate matter with diameter less than 2.5$\mu$m), which can be useful for epidemiological studies and for informing public health actions. One such study by \citet{Yao2018a} also used RF to estimate hourly concentrations of PM2.5 in British Columbia, Canada. \citet{Zou2019} compared RF, BRT and MLR to estimate regional PM2.5 concentrations in the Pacific Northwest and found RF performed much better than the other algorithms. 
In another very broad study covering several datasets and ML methods, \citet{Reid2015} estimated spatial distributions of PM2.5 concentrations during the 2008 northern California wildfires. The authors of the aforementioned study used 29 predictor variables and compared 11 different statistical models, including RF, BRT, SVM, and KNN. Overall, the BRT and RF models displayed the best performance. Emissions other than particulate matter have also been modelled using ML, as \citet{Lozhkin2016} used an ANN to predict carbon monoxide concentrations emitted from a peat fire in Siberia, Russia. \fixed{In another study, the authors used ten different statistical and ML methods and 21 covariates (including weather, geography, land-use, and atmospheric chemistry) to predict ozone exposures before and after wildfire events  \citep{Watson2019}. Here, gradient boosting gave the best results with respect to both root mean square error and $R^2$ values, followed by RF and SVM.}
In a different application related to smoke, \citet{Fuentes2019} used ANNs to detect smoke in several different grape varietals used for wine making. 

\subsubsection{Post-fire regeneration, succession, and ecology}

The study of post-fire regeneration is an important aspect of understanding forest and ecosystem responses and resilience to wildfire disturbances, with important ecological and economic consequences. RF, for example, has been a popular ML method for understanding the important variables driving post-fire regeneration \citep{Joao2018,Vijayakumar2016}. 
Burn severity (a measure of above and below ground biomass loss due to fire) is an important metric for understanding the impacts of wildfire on vegetation and post-fire regeneration, soils, and potential successional shifts in forest composition, and as such, has been included in many ML studies in this section, including \citep{Barrett2011,Cai2013,Cardil2019,Chapin2014,Divya2016,Fairman2017,Han2015,Johnstone2010,Liu2014a,Martin-Alcon2016,Sherrill2012,Thompson2010}. For instance, \citet{Cardil2019} used BRT to demonstrate that remotely-sensed data (i.e., Relative Differenced Normalized Burn Ratio index; RdNBR) can provide an acceptable assessment of fire-induced impacts (i.e., burn severity) on forest vegetation, while \citep{Fairman2017} used RF to identify the variables most important in explaining plot-level mortality and regeneration of Eucalyptus pauciflora in Victoria, Australia, affected by high-severity wildfires and subsequent re-burns. \citet{Debouk2013} assessed post-fire vegetation regeneration status using field measurements, a canopy height model, and Lidar (i.e., 3D laser scanning) data with a simple ANN.
	Post-fire regeneration also has important implications for the successional trajectories of forested areas, and a few studies have examined this using ML approaches \citep{Barrett2011,Cai2013,Johnstone2010}. For example, \citet{Barrett2011} used RF to model fire severity, from which they made an assessment of the area susceptible to a shift from coniferous to deciduous forest cover in the Alaskan boreal forest, while \citet{Cai2013} used BRT to assess the influence of environmental variables and burn severity on the composition and density of post-fire tree recruitment, and thus the trajectory of succession, in northeastern China.
	In other studies not directly related to post-fire regeneration, \citet{Hermosilla2015} used RF to attribute annual forest change to one of four categories, including wildfire, in Saskatchewan, Canada, while \citep{Jung2013} used GA and RF to estimate the basal area of post-fire residual spruce (\emph{Picea obovate}) and fir (\emph{Abies sibirica}) stands in central Siberia using remotely sensed data. \citet{Magadzire2019} used MaxEnt to demonstrate that fire return interval and species life history traits affected the distribution of plant species in South Africa. ML has also been used to examine fire effects on the hydrological cycle, as \citet{Poon2018} used SVM to estimate both pre- and post-wildfire evapotranspiration using remotely sensed variables.
	
	Considering the potential impacts of wildfires on wildlife, it is perhaps surprising that relatively few of such studies have adopted ML approaches. However, ML methods have been used to predict the impacts of wildfire and other drivers on species distributions and arthropod communities. \citet{Hradsky2017}, for example, used non-parametric BNs to describe and quantify the drivers of faunal distributions in wildfire-affected landscapes in southeastern Australia. Similarly, \citet{Reside2012} used MaxEnt to model bird species distributions in response to fire regime shifts in northern Australia, which is an important aspect of conservation planning in the region. ML has also been used to look at the effects of wildfire on fauna at the community level, as \citet{Luo2017} used DTs, Association Rule Mining, and AdaBoost to examine the effects of fire disturbance on spider communities in Cangshan Mountain, China.

\subsubsection{Socioeconomic effects}

ML methods have been little used to model socio-economic impacts of fire to date. 
We found one study in which BNs were used to predict the economic impacts of wildfires in Greece from 2006-2010 due to housing losses \citep{Papakosta2017}.  The authors did this by first defining a causal relationship between the participating variables, and then using BNs to estimate housing damages. It is worth noting that the problem of detecting these causal relationships from data is a difficult task and remains an active area of research in artificial intelligence.

\subsection{Fire management}

The goal of contemporary fire management is to have the appropriate amount of fire on the landscape, which may be accomplished through the management of vegetation including prescribed burning, the management of human activities (prevention), and fire suppression.   Fire management is a form of risk management that seeks to maximize fire benefits and minimize costs and losses \citep{Finney2005}. \fixed{Fire management decisions have a wide range of scales, including long-term strategic decisions about the acquisition and location of resources or the application of vegetation management in large regions, medium-term tactical decisions about the acquisition of additional resources, relocation, or release of resources during the fire season, and short-term real time operational decisions about the deployment and utilization of resources on individual incidents.} Fire preparedness and response is a supply chain with a hierarchical dependence. \citet{Taylor2020} describes 20 common decision types in fire management and maps the spatial-temporal dimensions of their decision spaces. 

Fire management models can be predictive, such as the probability of initial attack success, or prescriptive such as to maximize/minimize an objective function (e.g., optimal helicopter routing to minimize travel time in crew deployment).
 While advances have been made in the domain of wildfire management using ML techniques, there have been relatively few studies in this area compared to other wildfire problem domains. Thus, there appears to be great potential for ML to be applied to wildfire management problems, which may lead to novel and innovative approaches in the future.

\subsubsection{Planning and policy}

An important area of fire management is planning and policy, where various ML methods have been applied to address pertinent challenges. For example, \citet{Bao2015} used GA, which are useful for solving multi-objective optimization problems, to optimize watchtower locations for forest fire monitoring. \citet{Bradley2016} used RF to investigate the relationship between the protected status of forest in the western US and burn severity. Likewise, \citet{Ruffault2015} also used BRTs to assess the impact of fire policy introduced in the 1980s on fire activity in southern France and the relationships between fire and weather, and \citet{Penman2011} used BNs to build a framework to simultaneously assess the relative merits of multiple management strategies in Wollemi National Park, NSW, Australia. \citet{McGregor2016} used Markov decision processes (MDP) and model free Monte Carlo method to create fast running simulations (based on the FARSITE simulator) to create interactive visualizations of forest futures over 100 years based on alternate high-level suppression policies. \citet{McGregor2017} demonstrated ways in which a variety of ML and optimization methods can be used to create an interactive approximate simulation tool for fire managers. The authors of the aforementioned study utilized a modified version of the FARSITE fire-spread simulator, which was augmented to run thousands of simulation trajectories while also including new models of lightning strike occurrences, fire duration, and a forest vegetation simulator. \citet{McGregor2017} also clearly show how decision trees can be used to analyze a hierarchy of decision thresholds for deciding whether to suppress a fire or not; their hierarchy splits on fuel levels, then intensity estimations, and finally weather predictors to arrive at a generalizable policy. 

\subsubsection{Fuel treatment}
 
ML methods have also been used to model the effects of fuel treatments in order to mitigate wildfire risk. For example, \citet{Penman2014a} used a BN to examine the relative risk reduction of using prescribed burns on the landscape versus within the 500m interface zone adjacent to houses in the Sydney basin, Australia. \citet{Lauer2017} used approximate dynamic programming (also known as reinforcement learning) to determine the optimal timing and location of fuel treatments and timber harvest for a fire-threatened landscape in Oregon, USA, with the objective of maximizing wealth through timber management. Similarly, \citet{Arca2015} used GA for multi-objective optimization of fuel treatments.

\subsubsection{Wildfire preparedness and response}
 
Wildfire preparedness and response issues have also been examined using ML techniques. \citet{Costafreda-Aumedes2015} used ANNs to model the relationships between daily fire load, fire duration, fire type, fire size, and response time, as well as personnel and terrestrial/aerial units deployed for individual wildfires in Spain. Most of the models in \citet{Costafreda-Aumedes2015} highlighted the positive correlation of burned area and fire duration with the number of resources assigned to each fire, and some highlighted the negative influence of daily fire load. In another study, \citet{Penman2015} used Bayesian Networks to assess the relative influence of preventative and suppression management strategies on the probability of house loss in the Sydney basin, Australia. \citet{OConnor2017} used BRT to develop a predictive model of fire control locations in the Northern Rocky Mountains, USA, based on the likelihood of final fire perimeters, while \citet{Homchaudhuri2010} used GAs to optimize fireline generation. \citet{Rodrigues2019} modelled the probability that wildfire will escape initial attack using a RF model trained with fire location, detection time, arrival time, weather, fuel types, and available resources data. Important variables in \citet{Rodrigues2019} included fire weather and simultaneity of events. \citet{Julian2018} used two different RL algorithms to develop a system for autonomous control of one or more aircraft in order to monitor active wildfires. 

\subsubsection{Social factors}
Recently, the use of ML in fire management has grown to encompass more novel aspects of fire management, even including the investigation of criminal motives related to arson, as \citet{Delgado2018} used BNs to characterize wildfire arsonists in Spain thereby identifying five motivational archetypes (i.e., slight negligence; gross negligence; impulsive; profit; and revenge).

%% file: subdomains_methods_count_nice_gt5.tex
\begin{tabular}{lp{5cm}lllllllllllllr}
\toprule
Section &                                    Domain & NFM & SVM & KM &  GA & BN & BRT & ANN &  DT &  RF & KNN & MAXENT &  DL & NB & Other \\
\midrule
    1.1 &                    Fuels characterization &   - &   2 &  - &   - &  - &   1 &   1 &   1 &   4 &   1 &      - &   - &  - &     - \\
    1.2 &                            Fire detection &   2 &   3 &  1 &   1 &  1 &   - &  12 &   - &   - &   - &      - &  18 &  - &     3 \\
    1.3 &       Fire perimeter and severity mapping &   1 &  12 &  1 &   2 &  - &   1 &   6 &   1 &   4 &   2 &      1 &   - &  - &     6 \\
    2.1 &                   Fire weather prediction &   - &   - &  1 &   - &  - &   - &   - &   - &   1 &   - &      - &   - &  - &     3 \\
    2.2 &                      Lightning prediction &   - &   - &  - &   - &  - &   - &   - &   1 &   2 &   - &      - &   - &  - &     - \\
    2.3 &                            Climate change &   - &   1 &  - &   - &  - &   6 &   2 &   2 &   5 &   - &      7 &   - &  - &     - \\
    3.1 &                Fire occurrence prediction &   - &   3 &  - &   - &  1 &   - &   7 &   1 &   5 &   1 &      2 &   - &  1 &     4 \\
    3.2 &    Landscape-scale Burned area prediction &   - &   1 &  1 &   1 &  - &   - &   1 &   1 &   2 &   - &      1 &   1 &  - &     1 \\
    3.3 &               Fire Susceptibility Mapping &   2 &  12 &  1 &   3 &  2 &   8 &  16 &   9 &  26 &   - &     27 &   1 &  2 &     3 \\
    3.4 &                Landscape controls on fire &   2 &  10 &  1 &   3 &  2 &  19 &  11 &  15 &  40 &   1 &     30 &   1 &  1 &     2 \\
    4.1 &                    Fire Spread and Growth &   - &   - &  - &  13 &  2 &   - &   4 &   - &   1 &   1 &      - &   3 &  - &     2 \\
    4.2 &  Burned area and fire severity prediction &   - &   7 &  - &   1 &  1 &   3 &  10 &   7 &   6 &   3 &      - &   2 &  1 &     5 \\
    5.1 &                 Soil erosion and deposits &   - &   - &  1 &   - &  - &   - &   1 &   1 &   - &   - &      1 &   - &  - &     - \\
    5.2 &              Smoke and particulate levels &   - &   2 &  - &   - &  - &   3 &   3 &   - &   5 &   2 &      - &   - &  - &     2 \\
    5.3 &        Post-fire regeneration and ecology &   - &   1 &  - &   1 &  1 &   6 &   1 &   2 &  10 &   - &      2 &   - &  1 &     - \\
    5.4 &                     Socioeconomic effects &   - &   - &  - &   - &  1 &   - &   - &   - &   - &   - &      - &   - &  - &     - \\
    6.1 &                       Planning and policy &   - &   - &  - &   1 &  1 &   - &   - &   - &   2 &   - &      - &   - &  - &     2 \\
    6.2 &                            Fuel treatment &   - &   - &  - &   1 &  1 &   - &   - &   - &   - &   - &      - &   - &  - &     1 \\
    6.3 &        Wildfire preparedness and response &   - &   - &  - &   1 &  2 &   1 &   1 &   - &   - &   - &      1 &   1 &  - &     1 \\
    6.4 &                            Social factors &   - &   - &  - &   - &  1 &   - &   - &   - &   - &   - &      - &   - &  - &     - \\
\bottomrule
\end{tabular}

%% file: discussion_section.tex
\section{Discussion}\label{sec:discussion}

ML methods have seen a spectacular evolution in development, accuracy, computational efficiency, and application in many fields since the 1990s. It is therefore not surprising that ML has been helpful in providing new insights into several critical sustainability and social challenges in the 21st century \citep{gomes2009,Sullivan2014,Butler2017}. The recent uptake and success of ML methods has been driven in large part by ongoing advances in computational power and technology. For example, the recent use of bandwidth optimized Graphics Processing Units (GPUs) takes advantage of parallel processing for simultaneous execution of computationally expensive tasks, which has facilitated a wider use of computationally demanding but more accurate methods like DNNs. The advantages of powerful but efficient ML methods are therefore widely anticipated as being useful in wildfire science and management.

However, despite some early papers suggesting that data driven techniques would be useful in forest fire management \citep{Latham1987a, Kourtz1990, Kourtz1993}, our review has shown that there was relatively slow adoption of ML-based research in wildfire science up to the 2000s compared with other fields,  followed by a sharp increase in publication rate in the last decade.  In the early 2000s, data mining techniques were quite popular and classic ML methods such as DTs, RF, and bagging and boosting techniques began to appear in the wildfire science literature (e.g., \citet{Stojanova2006}). In fact, some researchers started using simple feed forward ANNs for small scale applications as early as the mid 1990s and early 2000s (e.g., \citet{Mccormick1999,Al-Rawi2002b}). In the last three decades, almost all major ML methods have been used in some way in wildfire applications, although some more computationally demanding methods, such as SOMs and cellular automatons, have only been actively experimented with in the last decade \citep{Toujani2018,Zheng2017}. Furthermore, the recent development of DL algorithms, with a particular focus on extracting spatial features from images, has led to a sharp rise in the application of DL for wildfire applications in the last decade. It is evident, however, from our review that while an increasing number of ML methodologies have been used across a variety of fire research domains over the past 30 years, this research is unevenly distributed among ML algorithms, research domains and tasks, and has had limited application in fire management.

Many fire science and management questions can be framed within a fire risk context. \citet{Xi2019} discussed the advantages of adopting a risk framework with regard to statistical modeling of wildfires. There the risk components of ``hazard’’, ``vulnerability’’ and ``exposure’’ are replaced respectively by fire probability, fire behavior and fire effects. Most fire management activities can be framed as risk controls to mitigate these components of risk. Traditionally, methods used in wildfire fire science to address these various questions have included physical modeling (e.g., \citet{Sullivan2009,Sullivan2009a,Sullivan2009b}), statistical methods (e.g., \citet{Taylor2013,Xi2019}), simulation modeling (e.g., \citet{Keane2004}), and operations research methods (\citet{Martell2015,Minas2012}).

In simple terms, any analytical study begins with one or more of four questions: ``what happened?’’; ``why did it happen?’’; ``what will happen?’’; or ``what to do?’’ Corresponding data driven approaches to address these questions are respectively called descriptive, diagnostic, predictive, and prescriptive analytics. The type of analytical approach adopted then circumscribes the types of methodological approaches (e.g., regression, classification, clustering, dimensionality reduction, decision making) and sets of possible algorithms appropriate to the analysis.

In our review, we found that studies incorporating ML methods in wildland fire science were predominantly associated with descriptive or diagnostic analytics, reflecting the large body of work on fire detection and mapping using classification methods, and on fire susceptibility mapping and landscape controls on fire using regression approaches. In many cases, the ML methods identified in our review are an alternative to statistical methods used for clustering and regression. While the aforementioned tasks are undoubtedly very important for understanding wildland fire, we found much less work associated with predictive or prescriptive analytics, such as fire occurrence prediction (predictive), fire behaviour prediction (predictive), and fire management (prescriptive). This may be because: a) particular domain knowledge is required to frame fire management problems; b) fire management data are often not publicly available, need a lot of work to transform into an easily analyzable form, or do not exist at the scale of the problem; and c) some fire management problems are not suited or can’t be fully addressed by ML approaches. We note that much of the work on fire risk in the fire susceptibility and mapping domain used historical fire and environmental data to map fire susceptibility; therefore, while that work aims to inform future fire risk, it cannot be considered to be predictive analytics, except, for example, in cases where it was used in combination with climate change projections. It appears then that, in general, wildfire science research is currently more closely aligned with descriptive and diagnostic analytics, whereas wildfire management goals are aligned with predictive and prescriptive analytics. This fundamental difference identifies new opportunities for research in fire management, which we discuss later in this paper. 

In the remainder of the paper, we examine some considerations for the use of ML methods, including: data considerations, model selection and accuracy, implementation challenges, interpretation, opportunities, and implications for fire management.

\subsection{Data considerations}

ML is a data-centric modeling paradigm concerned with finding patterns in data. Importantly, data scientists need to determine, often in collaboration with fire managers or domain experts, whether there are suitable and sufficient data for a given modeling task. Some of the criteria for suitable data include whether: a) the predictands and covariates are or can be wrangled into the same temporal and spatial scale; b) the observations are a representative sample of the full range of conditions that may occur in application of a model to future observations; and c) whether the data are at spatiotemporal scale appropriate to the fire science or management question. The first of these criteria can be relaxed in some ML models such as ANNs and DNNs, where inputs and outputs can be at different spatial or temporal scales for appropriately designed network architectures, although data normalization may still be required. The second criterion also addresses the important question of whether enough data exists for training a given algorithm for a given problem. In general, this question depends on the nature of the problem, complexity of the underlying model, data uncertainty and many other factors (see \citet{Roh2018} for a further discussion of data requirements for ML). In any case, many complex problems require a substantive data wrangling effort, to acquire, perform quality assurance, and fuse data into sampling units at the appropriate spatiotemporal scale. An example of this in daily fire occurrence prediction, where observations of a variety of features (e.g., continuous measures such as fire arrival time and location, or lightning strike times and locations) are discretized into three-dimensional (e.g., longitude, latitude, and day) cells called voxels. 
\fixed{Another important consideration for the collection and use of data in machine learning is selection bias. A form of spatial selection bias called preferential sampling occurs when sampling occurs preferentially in locations where one expects a certain response \citep{Diggle2010}. For example, preferential sampling may occur in air monitoring, because sensors may be placed in locations where poor air quality is expected \citep{Shaddick2014}. In general,  preferential sampling or other selection biases may be avoided altogether by selecting an appropriate sampling strategy at the experimental design phase, or, where this is not possible, to take it into account in model evaluation \citep{ Zadrozny2004}.}

For the problem domain fire detection and mapping, most applications of ML used some form of imagery (e.g., remote sensed satellite images or terrestrial photographs). In particular, many papers used satellite data (e.g., Landsat, MODIS) to determine vegetation differences before and after a fire and so were able to map area burned. For fire detection, many applications considered either remote sensed data for hotspot or smoke detection, or photographs of wildfires (used as inputs to an image classification problem). For fire weather and climate change, the three main sources of data were either weather station observations, climate reanalyses (modelled data that include historical observations), or GCMs for future climate projections. Reanalyses and GCMs are typically highly dimensional large gridded spatiotemporal datasets which require careful feature selection and/or dimensional reduction for ML applications. Fire occurrence prediction, susceptibility, and risk applications used a large number of different environmental variables as predictors, but almost all used fire locations and associated temporal information as predictands. Fire data itself is usually collated from fire management agencies in the form of georeferenced points or perimeter data, along with reported dates, ignition cause, and other related variables. Care should be taken using such data because changes in reporting standards or accuracy may lead to data inhomogeneity. As well as fire locations and perimeters, fire severity is an attribute of much interest to fire scientists. Fire severity is often determined from remotely sensed data and represented using variables such as the Differenced Normalized Burn Ratio (dNBR) and variants, or through field sampling. However, remote sensed estimates of burn severity should be considered as proxies as they have low skill in some ecosystems. Other fire ecology research historically relies on in situ field, sampling although many of the ML applications attempt to resolve features of interest using remote sensed data. Smoke data can also be derived from remote sensed imagery or from air quality sensors (e.g., PM2.5, atmospheric particulate matter less than 2.5 $\mu$m). 

Continued advances in remote sensing, as well as the quality and availability of remote sensed data products, in weather and climate modeling have led to increased availability of large spatiotemporal datasets, which presents both an opportunity and challenge for the application of ML methods in wildfire research and management. The era of ``big data'' has seen the development of cloud computing platforms to provide the computing and data storage facilities to deal with these large datasets. For example, in our review we found two papers \citep{Crowley2019,Quintero2019} that used Google Earth Engine which integrates geospatial datasets with a coding environment \citep{ gorelick2017google}. In any case, data processing and management plays an important role in the use of large geospatial datasets.

\subsection{Model selection and accuracy}

Given a wildfire science question or management problem and available relevant data, a critical question to ask is what is the most appropriate modeling tool to address the problem? Is it a standard statistical model (e.g., linear regression or LR), a physical model (e.g., FIRETEC or other fire simulator), a ML model, or a combination of approaches? Moreover, which specific algorithm will yield the most accurate classification or regression. Given the heterogeneity of research questions, study areas, and datasets considered in the papers reviewed here, it is not possible to comprehensively answer these questions with respect to ML approaches. Even in the case where multiple studies used the same dataset 
\citep{Cortez2007,Safi2013, Storer2016,Castelli2015a, AlJanabi2018,Alberg2015, Li2018,Castelli2015a}
the different research questions considered meant a direct comparison of ML methods was not possible between research studies. However, a number of individual studies did make comparisons between multiple ML methods, or between ML and statistical methods for a given wildfire modeling problem and dataset. Here we highlight some of their findings to provide some guidance with respect to model selection. In our review (see section \ref{sec:domains} and \fixed{the supplementary material}), we found \fixed{29} papers comparing ML and statistical methods, where in the majority of these cases ML methods were found to be more accurate than traditional statistical methods (e.g., GLMs), or displayed similar performance \citep{Pu2004,Bates2017,deBem2019}. In only one study on climate change by \citet{Amatulli2013}, MARS was found to be superior to RF for their analytical task. A sizable number of the comparative studies (14) involved classification problems that used LR as a benchmark method against ANN or ensemble tree methods. For studies comparing multiple ML methods, there was considerable variation in the choice of most accurate method; however, in general ensemble methods tended to outperform single classifier methods \fixed{(e.g., \citet{Stojanova2012,Dutta2016,Mayr2018,Nelson2017,Reid2015,Watson2019})}, except in one case where the most accurate model (CART) was also the most parsimonious \citep{ Coffield2019}. A few more recent papers also highlighted the advantages of DL over other methods. In particular, for fire detection, \citet{Zhang2018a} compared CNNs with SVM and found that CNNs were more accurate, while \citet{Zhao2018} similarly found CNNs superior to SVMs and ANNs. For fire susceptibility mapping, \citet{Zhang2019} found CNNs were more accurate than RF, SVMs, and ANNs. For time series forecasting problems, \citet{ Liang2019} found LSTMs outperformed ANNs. Finally, \citet{Cao2019} found that using an LSTM combined with a CNN led to better fire detection performance from video compared with CNNs alone.

In any case, more rigorous inter-model comparisons are needed to reveal in which conditions, and in what sense particular methods are more accurate, as well as to establish procedures for evaluating accuracy. ML methods are also prone to overfitting, so it is important to evaluate \fixed{models} with robust test datasets using appropriate cross-validation strategies. \fixed{For example, the na\"ive application of cross-validation to data that have spatial or  spatio-temporal dependencies may lead to overly optimistic evaluations \citep{Roberts2017}.} In general, one \fixed{also} desires to minimise errors associated with either under-specification or over-specification of the model, a problem known as the bias-variance trade-off \citep{Geman1992}. However, several recent advances have been made to reduce overfitting in ML models, for instance, regularization techniques in DNNs \citep{Kukacka2017}. Moreover, when interpreting comparisons between ML and statistical methods, we should be cognizant that just as some ML methods require expert knowledge, the accuracy of statistical methods can also vary with the skill of the practitioner. \citet{Thompson2011a} also emphasize the need for identifying sources of uncertainty in modeling so that they can better managed.

\subsection{Implementation Challenges}

Beyond data and model selection, two important considerations for model specification are feature selection and spatial autocorrelation. Knowledge of the problem domain is extremely important in identifying a set of candidate features. However, while many ML methods are not limited by the number of features, more variables do not necessarily make for a more accurate, interpretable, or easily implemented model \citep{Schoenberg2016,Breiman2001} and can lead to overfitting and increased computational time. Two different ML methods to enable selection of a reduced and more optimal set of features include GAs and PSO. \citet{Sachdeva2018} used a GA to select input features for BRT and found this method gave the best accuracy compared with ANN, RF, SVM, SVM with PSO (PSO-SVM), DTs, logistic regression, and NB. \citet{Hong2018} employed a similar approach for fire susceptibility mapping and found this led to improvements for both SVM and RF compared with their non-optimized counterparts. \citet{Tracy2018} used a novel random subset feature selection algorithm for feature selection, which they found led to higher AUC values and lower model complexity. \citet{Jaafari2019a} used a NFM combined with the imperialist competitive algorithm (a variant of GA) for feature selection which led to very high model accuracy (0.99) in their study. \citet{TienBui2017} used PSO to choose inputs to a NFN and found this improved results. \citep{Zhang2019} also considered the information gain ratio for feature selection. As noted in \citet{Moritz2012} and \citet{Mayr2018}, one should also take spatial autocorrelation into account when modeling fire probabilities spatially. In general, the presence of spatial autocorrelation violates the assumption of independence for parametric models, which can degrade model performance. One approach to deal with autocorrelation requires subsampling to remove any spatial autocorrelation \citet{Moritz2012}. It is also often necessary to subsample from non-fire locations due to class imbalance between ignitions and non-ignitions (e.g., \citet{Cao2017, Zhang2019}). \citet{Song2017} considered spatial econometric models and found a spatial autocorrelation model worked better than RF, although \citet{Kim2019} note that RF may be robust to spatial autocorrelation with large samples. In contrast to many ML methods, a strength of CNNs is its ability to exploit spatial correlation in the data to enable the extraction of spatial features.

\subsection{Interpretation}

A major obstacle for the adoption of ML methods to fire modeling tasks is the perceived lack of interpretability or explainability of such methods, which are often considered to be ``black box’’ models. Users (in this case fire fighters and managers) need to trust ML model predictions, and so have the confidence and justification to apply these models, particularly in cases where proposed solutions are considered novel. Model intepretability should therefore be an important aspect of model development if models are to be selected and deployed in fire management operations. Model interpretability varies significantly across the different types of ML. For example, conventional thinking is that tree-based methods are more interpretable than neural network methods. This is because a single decision tree classifier can be rendered as a flow chart corresponding to if-then-else statements, whereas an ANN represents a nonlinear function approximated through a series of nonlinear activations. However, because they combine multiple trees in an optimized way, ensemble tree classifiers are less interpretable than single tree classifiers. On the other hand, BNs are one example of an ML technique where good explanations for results can be inferred due to their graphical representation; however, full Bayesian learning on large-scale data is very computationally expensive which may have limited early applications; however, as computational power has increased we have seen an increase in the popularity of BNs in wildfire science and management applications (e.g., \citet{ Penman2015,Papakosta2017}).

DL-based architectures are widely considered to be among the least interpretable ML models, despite the fact that they can achieve very accurate function approximation \citep{Chakraborty2017}. In fact, this is demonstrative of the well-known trade-off between prediction accuracy and interpretability (see \citet{Kuhn2013} for an in-depth discussion). The ML community, however, recognizes the problem of interpretability and work is underway to develop methods that allow for greater interpretability of ML methods, including methods for DL (see for example, \citet{McGovern2019}) or model-agnostic approaches \citep{Ribeiro2016}. \citet{Runge2019} further argue that casual inference methods should be used in conjunction with predictive models to improve our understanding of physical systems. Finally, it is worth noting that assessing variable importance (see Sec. \ref{sec:controls}) for a given model can play a role in model interpretation.

\subsection{Opportunities}

Our review highlights a number of potential opportunities in wildfire science and management for ML applications where ML has not yet been applied or is under-utilized. Here we examine ML advances in other areas of environmental science that have analogous problems in wildland fire science and which may be useful for identifying further ML applications. For instance, \citet{Li2011b} compared ML algorithms for spatial interpolation and found that a RF model combined with geostatistical methods yielded good results; a similar method could be used to improve interpolation of fire weather observations from weather stations, and so enhance fire danger monitoring. \citet{Rasp2018} showed that ANNs could improve weather forecasts by post-processing ensemble forecasts, an approach which could similarly be applied to improve short-term forecasts of fire weather. \citet{Belayneh2014a} used ANNs and SVMs combined with wavelet transforms for long term drought forecasting in Ethiopia; such methods could also be useful for forecasting drought in the context of fire danger potential. In the context of numerical weather prediction, \citet{Cohen2019} found better predictability using ML methods than dynamical models for subseasonal to seasonal weather forecasting, suggesting similar applications for long-term fire weather forecasting. \citet{McGovern2017} discussed how AI techniques can be leveraged to improve decision making around high-impact weather. More recently, \citet{Reichstein2019} have further argued for the use of DL in the environmental sciences, citing its potential to extract spatiotemporal features from large geospatial datasets. \citet{Kussul2017} used CNNs to classify land cover and crop types and found that CNNs improved the results over standard ANN models; a similar approach could be used for fuels classification, which is an important input to fire behaviour prediction models. \citet{Shi2016} also used CNNs to detect clouds in remote sensed imagery and were able to differentiate between thin and thick cloud. A similar approach could be used for smoke detection, which is important for fire detection, as well as in determining the presence of false negatives in hotspot data (due to smoke or cloud obscuration). Finally, recent proposals have called for hybrid models that combine process-based models and ML methods \citep{Reichstein2019}. For example, ML models may replace user-specified parameterizations in numerical weather prediction models \citep{Brenowitz2018}. Other recent approaches use ML methods to determine the solutions to nonlinear partial differential equations \citet{Raissi2018, Raissi2019}. Such methods could find future applications in improving fire behaviour prediction models based on computationally expensive physics-based fire simulators, in coupled fire-atmosphere models, or in smoke dispersion modeling. In any case, the applications of ML that we have outlined are meant for illustrative purposes and are not meant to represent an exhaustive list of all possible applications.

\subsection{Implications for fire management}

We believe ML has been under-utilized in fire management, particularly with respect to problems belonging to either predictive or prescriptive analytics. Fire management comprises a set of risk control measures, which are often cast in the framework of the emergency response phases: prevention; mitigation; preparedness; response; recovery; and review \citep{Tymstra2019}. In terms of financial expenditure, by far the largest percentage spent in the response phase \citep{Stocks2016}. In practice, fire management is largely determined by the need to manage resources in response to active or expected wildfires, typically for lead times of days to weeks, or to manage vegetative fuels. This suggests the opportunity for increased research in areas of fire weather prediction, fire occurrence prediction, and fire behaviour prediction, as well as optimizing fire operations and fuel treatments. The identification of these areas, as well as the fact that wildfire is both a spatial and temporal process, further reiterate the need for ML applications for time series forecasting. 

From this review, there were few papers that used time series ML methods for forecasting problems, suggesting an opportunity for further work in this area. In particular, recurrent neural networks (RNNs) were used for fire behavior prediction \citep{Cheng2008, Kozik2013,Kozik2014} and fire occurrence prediction \citep{Dutta2013}. The most common variant of RNNs are Long Short Term Memory (LSTM) networks \citep{hochreiter1997long}, which have been used for burned area prediction \citep{Liang2019} and fire detection \citep{Cao2019}. Because these methods implicitly model dynamical processes, they should lead to improve forecasting models compared with standard ANNs. For example \citet{Gensler2017} have used LSTMs to forecast solar power and \citet{Kim2017} used CNNs combined with LSTM for forecasting precipitation. We anticipate that these methods could also be employed for fire weather, fire occurrence, and fire behaviour prediction.

We note that there are a number of operational research and management science methods used in fire management research including queuing, optimization, and simulation of complex system dynamics (e.g., \citet{Martell2015}) where ML algorithms don’t seem to provide an obvious alternative. For example, planning models to simulate the interactions between fire management resource configurations and fire dynamics reviewed by \citep{Mavsar2013}. From our review, a few papers used agent-based learning methods for fire management. In particular, reinforcement learning was used for optimizing fuel treatments \citep{Lauer2017} or for autonomous control of aircraft for fire monitoring \citep{Julian2018}. GAs were used for generating optimal firelines for active fires \citep{ Homchaudhuri2010} and for reducing the time for fire simulation \citep{Cencerrado2014}. However, more work is needed to identify where ML methods could contribute to tactical, operational, or strategic fire management decision making. 

An important challenge for the fire research and management communities is enabling the transition of potentially useful ML models to fire management operations. Although we identified several papers that emphasized their ML models could be deployed in fire management operations \citep{Artes2016,Alonso-Betanzos2002,Iliadis2005,Stojanova2012,Davis1989,Davis1986,Liu2015}, it can be difficult to assess whether and how a study has been adopted by, or influenced, fire management agencies. This challenge is often exacerbated by a lack of resources and/or funding, as well as the different priorities and institutional cultures of researchers and fire managers. One possible solution to this problem would be the formation of working groups dedicated to enabling this transition, preferably at the research proposal phase. In general, enabling operational ML methods will require tighter integration and greater collaboration between the research and management communities, particularly with regards to project design, data compilation and variable selection, implementation, and interpretation. However, it is worth noting that this is not a problem unique to ML, it is a long-standing and common issue in many areas of fire research and other applied science disciplines, where continuous effort is required to maintain communications and relationships between researchers and practitioners.

Finally, we would like to stress that we believe the wildfire research and management communities should play an active role in providing relevant, high quality, and freely available wildfire data for use by practitioners of ML methods. For example, burned area and fire weather data made available by \citet{Cortez2007} was subsequently used by a number of authors in their work. It is imperative that the quality of data collected by management agencies be as robust as possible, as the results of any modelling process are dependent upon the data used for analysis. It is worth considering how new data on, for example, hourly fire growth or the daily use of fire management resources, could be used in ML methods to yield better predictions or management recommendations — using new tools to answer new questions may require better or more complete data. Conversely, we must recognize that despite ML models being able to learn on their own, expertise in wildfire science is necessary to ensure realistic modelling of wildfire processes, while the complexity of some ML methods (e.g., DL) requires a dedicated and sophisticated knowledge of their application (we note that many of the most popular ML methods used in this study are fairly easy to implement, such as RF, MaxEnt, and DTs). The observation that no single ML algorithm is superior for all classes of problem, an idea encapsulated by the ``no free lunch’’ theorem \citep{Wolpert1996}, further reinforces the need for domain-specific knowledge. Thus, the proper implementation of ML in wildfire science is a challenging endeavor, often requiring multidisciplinary teams and/or interdisciplinary specialists to effectively produce meaningful results.

\subsection{A word of caution}

ML holds tremendous potential for a number of wildfire science and management problem domains. As indicated in this review, much work has already been undertaken in a number of areas, although further work is clearly needed for fire management specific problems. Despite this potential, ML should not be considered a panacea for all fire research areas. ML is best suited to problems where there is sufficient high-quality data, and this is not always the case. For example, for problems related to fire management policy, data is needed at large spatiotemporal scales (i.e., ecosystem/administrative spatial units at timescales of decades or even centuries), and such data may simply not yet exist in current inventories. At the other extreme, data is needed at very fine spatiotemporal scales for fire spread and behavior modeling, including high resolution fuel maps and surface weather variables which are often not available at the required scale and are difficult to acquire even in an experimental context. Another limitation of ML may occur when one attempts make predictions where no analog exists in the observed data, such as may be the case with climate change prediction.

%% file: conclusions_section.tex
\section{Conclusions}\label{sec:conclusions}

Our review shows that the application of ML methods in wildfire science and management has been steadily increasing since their first use in the 1990s, across core problem domains using a wide range ML methods. The bulk of work undertaken thus far has used traditional methods such as RF, BRT, MaxEnt, SVM and ANNs, partly due to the ease of application and partly due to their simple interpretability in many cases. However, problem domains associated with predictive (e.g., predicted fire behavior) or prescriptive analytics (e.g. optimizing fire management decisions) have seen much less work with ML methods. We therefore suggest opportunities exist for both the wildfire community and ML practitioners to apply ML methods in these areas. Moreover, the increasing availability of large spatio-temporal datasets, from climate models or remote sensing for example, may be amenable to the use of deep learning methods, which can efficiently extract spatial or temporal features from data. Another major opportunity is the application of agent based learning to fire management operations, although many other opportunities exist. However, we must recognize that despite ML models being able to learn on their own, expertise in wildfire science is necessary to ensure realistic modelling of wildfire processes across multiple scales, while the complexity of some ML methods (e.g. DL) requires a dedicated and sophisticated knowledge of their application. Furthermore, a major obstacle for the adoption of ML methods to fire modeling tasks is the perceived lack of interpretability of such methods, which are often considered to be black box models. The ML community, however, recognizes this problem and work is underway to develop methods that allow for greater interpretability of ML methods (see for example, \citep{McGovern2019}). 
Data driven approaches are by definition data dependent  --- if the fire management community wants to more fully exploit  powerful ML methods, we need to consider data as a valuable resource and examine what further information on fire events or operations are needed to apply ML approaches to management problems.  
Thus, wildland fire science is a diverse multi-faceted discipline that requires a multi-pronged approach, a challenge made greater by the need to mitigate and adapt to \fixed{current and future fire regimes}.


%% file: ML-Fire-Paper-Table-Simple.tex
\section*{S.1. Fuels Characterization, Fire Detection And Mapping}

\subsection*{S.1.1 Fuels characterization}

\begin{longtable}{p{4cm}p{4.5cm}p{6cm}}
\toprule
                  Citation & ML methods used &                                    Study Region \\
\midrule
\endhead
\midrule
\multicolumn{3}{r}{{Continued on next page}} \\
\midrule
\endfoot

\bottomrule
\endlastfoot
         \citet{Riano2005} &             ANN &                                   Not specified \\
        \citet{Garcia2011} &             SVM &           Alto Tajo Natural Park, central Spain \\
        \citet{Pierce2012} &              RF &  Lassen Volcanic National Park, California, USA \\
       \citet{Chirici2013} &     DT, RF, BRT &                                   Sicily, Italy \\
         \citet{Riley2014} &              RF &                             Eastern Oregon, USA \\
 \citet{Lopez-Serrano2016} &    SVM, KNN, RF &                 Sierra Madre Occidental, Mexico \\
\end{longtable}

\subsection*{S.1.2 Fire detection}

\begin{longtable}{p{4cm}p{4.5cm}p{6cm}}
\toprule
                 Citation &              ML methods used &                                       Study Region \\
\midrule
\endhead
\midrule
\multicolumn{3}{r}{{Continued on next page}} \\
\midrule
\endfoot

\bottomrule
\endlastfoot
        \citet{Arrue2000} &                          ANN &               Experiments at University of Seville \\
      \citet{Al-Rawi2001} &                          ANN &                                      Eastern Spain \\
   \citet{ZhanqingLi2001} &                          ANN &                                             Canada \\
        \citet{Utkin2002} &                          ANN &                                      Not specified \\
      \citet{Cordoba2004} &                           GA &                                        Experiments \\
    \citet{Fernandes2004} &                          ANN &                                      Not specified \\
   \citet{Fernandes2004a} &                          ANN &                                      Not specified \\
    \citet{Srinivasa2008} &                           KM &                                      Not specified \\
 \citet{Angayarkkani2010} &                          ANN &                                      Not specified \\
           \citet{Ko2010} &                           BN &                                        test images \\
      \citet{Soliman2010} &                          ANN &                             Laboratory experiments \\
 \citet{Angayarkkani2011} &                        ANFIS &                                      not specified \\
         \citet{Wang2011} &                        ANFIS &                                              Tibet \\
         \citet{Zhao2011} &                     SVM, GMM &                                        Test images \\
           \citet{Li2015} &                          ANN &  China, North East Asia, Russia, Canada, Australia \\
          \citet{Liu2015} &                          ANN &                             Laboratory experiments \\
        \citet{Zhang2016} &                     CNN, SVM &                                        Test images \\
     \citet{Akhloufi2018} &                          CNN &                                            Corsica \\
          \citet{Li2018a} &                          CNN &                                        Test images \\
     \citet{Muhammad2018} &                          CNN &                                        Test images \\
         \citet{Yuan2018} &                          CNN &                                        Test images \\
        \citet{Zhang2018} &                          CNN &                                        Test images \\
       \citet{Zhang2018a} &                          CNN &                                     Synthetic data \\
         \citet{Zhao2018} &                          CNN &                                        Test images \\
   \citet{Alexandrov2019} &  CNN, HAAR{\_}CASCADES, YOLO &                                        Test images \\
           \citet{Ba2019} &                          CNN &                              Satellite test images \\
   \citet{Barmpoutis2019} &                          CNN &                                        Test images \\
          \citet{Cao2019} &                    CNN, LSTM &                              Test images and video \\
      \citet{Hossain2019} &                          ANN &                                        Test images \\
   \citet{Jakubowski2019} &                          CNN &                                        Test images \\
    \citet{JoaoSousa2019} &                          CNN &                                            Corsica \\
           \citet{Li2019} &                          CNN &                                        Test images \\
         \citet{Phan2019} &                          CNN &                                 American Continent \\
        \citet{Sayad2019} &                     ANN, SVM &                                             Canada \\
         \citet{Wang2019} &                          CNN &                                        Test images \\
\end{longtable}

\subsection*{S.1.3 Fire perimeter and severity mapping}

\begin{longtable}{p{4cm}p{4.5cm}p{6cm}}
\toprule
                  Citation &          ML methods used &                                       Study Region \\
\midrule
\endhead
\midrule
\multicolumn{3}{r}{{Continued on next page}} \\
\midrule
\endfoot

\bottomrule
\endlastfoot
       \citet{Al-Rawi2001} &                      ANN &                                      Eastern Spain \\
        \citet{Brumby2001} &                       GA &                 Cerro Grande Fire, New Mexico, USA \\
         \citet{Sunar2001} &             ANN, ISODATA &                              south coast of Turkey \\
      \citet{Al-Rawi2002b} &                      ANN &                                    Valencia, Spain \\
            \citet{Sa2003} &                  DT, BAG &                                Northern Mozambique \\
            \citet{Pu2004} &                      ANN &                           Northern California, USA \\
        \citet{Zammit2006} &             SVM, KM, KNN &                                    Southern France \\
 \citet{Alonso-Benito2008} &                      SVM &                          Tenerife and Gran Canaria \\
           \citet{Cao2009} &                      SVM &                                 Mongolia and China \\
         \citet{Celik2010} &                       GA &                  Reno Lake Tahoe area, Nevada, USA \\
   \citet{Petropoulos2010} &                      SVM &                                             Greece \\
       \citet{Dragozi2011} &                 SVM, KNN &                                      Stresa, Italy \\
         \citet{Gomez2011} &                      ANN &                                  Iberian Peninsula \\
   \citet{Petropoulos2011} &                      SVM &                                             Greece \\
      \citet{Mitrakis2012} &  NFM, ANN, SVM, ADABOOST &                                             Greece \\
       \citet{Dragozi2014} &                      SVM &                        Parnitha and Rhodes, Greece \\
    \citet{Hultquist2014a} &              GP, RF, SVM &                           Big Sur, California, USA \\
          \citet{Zhao2015} &                      SVM &                 Greater Yellowstone Ecosystem, USA \\
      \citet{Hamilton2017} &                      SVM &                                         Idaho, USA \\
      \citet{Hawbaker2017} &                      BRT &                                                USA \\
       \citet{Pereira2017} &                      SVM &                            Cerrado savanna, Brazil \\
       \citet{Collins2018} &                       RF &                                Victoria, Australia \\
         \citet{Nitze2018} &                       RF &  Alaska, Eastern Canada, Western Siberia and Ea... \\
       \citet{Crowley2019} &                     BULC &       Elephant Hill Fire, British Columbia, Canada \\
      \citet{Langford2019} &                      DNN &                               Interior Alaska, USA \\
      \citet{Quintano2019} &                   MAXENT &                                  La Cabrera, Spain \\
      \citet{Quintero2019} &                       RF &                                 West Central Spain \\
\end{longtable}

\section*{S.2. Fire Weather And Climate Change}

\subsection*{S.2.1 Fire weather prediction}

\begin{longtable}{p{4cm}p{4.5cm}p{6cm}}
\toprule
               Citation & ML methods used &     Study Region \\
\midrule
\endhead
\midrule
\multicolumn{3}{r}{{Continued on next page}} \\
\midrule
\endfoot

\bottomrule
\endlastfoot
    \citet{Skinner2002} &              KM &           Canada \\
   \citet{Crimmins2006} &             SOM &    Southwest USA \\
   \citet{Sanabria2013} &              RF &        Australia \\
 \citet{Lagerquist2017} &             SOM &  Alberta, Canada \\
    \citet{Nauslar2019} &             SOM &    Southwest USA \\
\end{longtable}

\subsection*{S.2.2 Lightning prediction}

\begin{longtable}{p{4cm}p{4.5cm}p{6cm}}
\toprule
           Citation & ML methods used &     Study Region \\
\midrule
\endhead
\midrule
\multicolumn{3}{r}{{Continued on next page}} \\
\midrule
\endfoot

\bottomrule
\endlastfoot
 \citet{Blouin2016} &              RF &  Alberta, Canada \\
  \citet{Bates2017} &          DT, RF &        Australia \\
\end{longtable}

\subsection*{S.2.3 Climate change}

\begin{longtable}{p{4cm}p{4.5cm}p{6cm}}
\toprule
               Citation &                ML methods used &                      Study Region \\
\midrule
\endhead
\midrule
\multicolumn{3}{r}{{Continued on next page}} \\
\midrule
\endfoot

\bottomrule
\endlastfoot
     \citet{Moritz2012} &                         MAXENT &                            Global \\
   \citet{Amatulli2013} &                             RF &              Mediterranean Europe \\
   \citet{Batllori2013} &                         MAXENT &   Mediterranen ecosystems, Global \\
       \citet{Liu2016a} &                        BRT, RF &                       Western USA \\
      \citet{Parks2016} &                            BRT &                       Western USA \\
 \citet{VanBreugel2016} &  RF, SVM, BRT, MAXENT, ANN, DT &                          Ethiopia \\
      \citet{Davis2017} &                         MAXENT &            Pacific Northwest, USA \\
         \citet{Li2017} &                         MAXENT &  Yunnan Province, Southwest China \\
      \citet{Young2017} &                            BRT &                       Alaska, USA \\
  \citet{Boulanger2018} &                    RF, BRT, DT &                            Canada \\
 \citet{Stralberg2018a} &                             RF &                   Alberta, Canada \\
      \citet{Stroh2018} &                         MAXENT &                 South central USA \\
      \citet{Tracy2018} &                         MAXENT &             Western North America \\
   \citet{Buckland2019} &                            ANN &                     Nebraska, USA \\
      \citet{Young2019} &                            BRT &                       Alaska, USA \\
\end{longtable}

\section*{S.3. Fire Occurrence, Susceptibility and Risk}

\subsection*{S.3.1 Fire occurrence prediction}

\begin{longtable}{p{4cm}p{4.5cm}p{6cm}}
\toprule
                    Citation &                          ML methods used &                                Study Region \\
\midrule
\endhead
\midrule
\multicolumn{3}{r}{{Continued on next page}} \\
\midrule
\endfoot

\bottomrule
\endlastfoot
     \citet{Vega-Garcia1996} &                                      ANN &                             Alberta, Canada \\
 \citet{Alonso-Betanzos2002} &                                      ANN &                    Galicia, Northwest Spain \\
 \citet{Alonso-Betanzos2003} &                                      ANN &                    Galicia, Northwest Spain \\
       \citet{Vasilakos2007} &                                      ANN &                       Lesvos Island, Greece \\
            \citet{Sakr2010} &                                      SVM &                                     Lebanon \\
            \citet{Sakr2011} &                                 SVM, ANN &                                     Lebanon \\
       \citet{Stojanova2012} &  KNN, NB, DT, SVM, BN, ADABOOST, BAG, RF &                                    Slovenia \\
           \citet{Dutta2013} &                                 ANN, DNN &                                   Australia \\
            \citet{Chen2015} &                                   MAXENT &  Daxinganling Mountains, Northeastern China \\
       \citet{DeAngelis2015} &                                   MAXENT &                  Canton Ticino, Switzerland \\
           \citet{Dutta2016} &                                      DNN &                                   Australia \\
     \citet{Vecin-Arias2016} &                                       RF &            Central Iberian Peninsula, Spain \\
             \citet{Cao2017} &                                  ANN, RF &                      Yunnan Province, China \\
              \citet{Yu2017} &                                       RF &                                    Cambodia \\
     \citet{VanBeusekom2018} &                                       RF &                                 Puerto Rico \\
\end{longtable}

\subsection*{S.3.2 Landscape-scale Burned area prediction}

\begin{longtable}{p{4cm}p{4.5cm}p{6cm}}
\toprule
              Citation &  ML methods used &              Study Region \\
\midrule
\endhead
\midrule
\multicolumn{3}{r}{{Continued on next page}} \\
\midrule
\endfoot

\bottomrule
\endlastfoot
     \citet{Cheng2008} &              RNN &                    Canada \\
 \citet{Archibald2009} &               RF &           Southern Africa \\
    \citet{Arnold2014} &      HCL, MAXENT &      Interior Western USA \\
      \citet{Mayr2018} &  DT, RF, SVM, KM &                   Namibia \\
     \citet{deBem2019} &          ANN, GA &  Federal District, Brazil \\
\end{longtable}

\subsection*{S.3.3 Fire Susceptibility Mapping}

\begin{longtable}{p{4cm}p{4.5cm}p{6cm}}
\toprule
                             Citation &                     ML methods used &                                  Study Region \\
\midrule
\endhead
\midrule
\multicolumn{3}{r}{{Continued on next page}} \\
\midrule
\endfoot

\bottomrule
\endlastfoot
                 \citet{Chuvieco1999} &                                 ANN &                          Mediterranean Europe \\
 \citet{PerestrelloDeVasconcelos2001} &                                 ANN &                              central Portugal \\
                 \citet{Amatulli2006} &                                  DT &                      Gargano Peninsula, Italy \\
                 \citet{Amatulli2007} &                                  DT &                                Tuscany, Italy \\
                   \citet{Lozano2008} &                                  DT &                            Northwestern Spain \\
                   \citet{Holden2009} &                                  RF &         Gila National Forest, New Mexico, USA \\
                    \citet{Maeda2009} &                                 ANN &                                        Brazil \\
                 \citet{Mallinis2009} &                              DT, KM &                                        Greece \\
                 \citet{Parisien2009} &                         MAXENT, BRT &                                           USA \\
                  \citet{Barrett2011} &                                  RF &                                        Alaska \\
                 \citet{Dimuccio2011} &                                 ANN &                              Central Portugal \\
                  \citet{Dlamini2011} &                                  BN &                                     Swaziland \\
                 \citet{Bisquert2012} &                                 ANN &                      Galicia, Northwest Spain \\
                   \citet{Moritz2012} &                              MAXENT &                                        Global \\
                 \citet{Oliveira2012} &                                  RF &                          Mediterranean Europe \\
                 \citet{Parisien2012} &                              MAXENT &                                   Western USA \\
                   \citet{Renard2012} &                              MAXENT &                          Western Ghats, India \\
                  \citet{Syphard2012} &                              MAXENT &                      Southern California, USA \\
               \citet{BarMassada2013} &                          RF, MAXENT &                                 Michigan, USA \\
                      \citet{Luo2013} &                                  RF &                                        global \\
                   \citet{Peters2013} &                              MAXENT &                                 Northeast USA \\
                  \citet{Syphard2013} &                              MAXENT &  South coast ecoregion, San Diego County, USA \\
                   \citet{Arpaci2014} &                          RF, MAXENT &                          Tyrol, European Alps \\
                 \citet{Parisien2014} &                                  DT &                                        Canada \\
               \citet{Rodrigues2014a} &                        RF, BRT, SVM &                              Peninsular Spain \\
                    \citet{Duane2015} &                              MAXENT &                              Catalonia, Spain \\
                  \citet{Bashari2016} &                                  BN &                        Isfahan province, Iran \\
                     \citet{Curt2016} &                                 BRT &                           Southeastern France \\
                  \citet{Fonseca2016} &                              MAXENT &                                        Brazil \\
                 \citet{Goldarag2016} &                                 ANN &              Golestan province, Northern Iran \\
                      \citet{Guo2016} &                                  RF &          Daxing'an Mountains, Northeast China \\
                     \citet{Guo2016a} &                                  RF &                        Fujian province, China \\
                \citet{Pourtaghi2016} &                             BRT, RF &              Golestan province, Northern Iran \\
                    \citet{Satir2016} &                                 ANN &                    Upper Seyhan Basin, Turkey \\
               \citet{VanBreugel2016} &       RF, SVM, BRT, MAXENT, ANN, DT &                                      Ethiopia \\
                    \citet{Vilar2016} &                              MAXENT &                          Madrid region, Spain \\
                     \citet{Adab2017} &                                 ANN &                                Northeast Iran \\
                      \citet{Cao2017} &                             ANN, RF &                        Yunnan Province, China \\
                    \citet{Davis2017} &                              MAXENT &                        Pacific Northwest, USA \\
                 \citet{Ebrahimy2017} &                              MAXENT &                            Eastern Azerbaijan \\
                       \citet{Li2017} &                              MAXENT &              Yunnan Province, Southwest China \\
                  \citet{Mostafa2017} &                                 SVM &              Golestan province, Northern Iran \\
                   \citet{Peters2017} &                              MAXENT &                                 Northeast USA \\
                     \citet{Song2017} &                                  RF &                             Hefei City, China \\
                  \citet{TienBui2017} &                   NFM, PSO, RF, SVM &                    Lam Dong province, Vietnam \\
                   \citet{Valdez2017} &                                  RF &                                      Honduras \\
                     \citet{Adab2018} &                              MAXENT &                     Mazandaran province, Iran \\
                     \citet{Hong2018} &                         SVM, RF, GA &                    Southwest Jiangxi Province \\
                  \citet{Jaafari2018} &                          DT, DT, NB &                        Zagros Mountains, Iran \\
                    \citet{Kahiu2018} &                                 BRT &                            sub Saharan Africa \\
              \citet{Leuenberger2018} &                             RF, ANN &                         Dao, Lafoes, Portugal \\
                \citet{NgocThach2018} &                        SVM, RF, ANN &                  Thuan Chau district, Vietnam \\
                    \citet{Parks2018} &                                 BRT &                                   Western USA \\
                 \citet{Sachdeva2018} &  BRT, ANN, RF, SVM, DT, NB, GA, PSO &                                Northern India \\
                 \citet{Tehrany2018a} &                         LB, SVM, RF &                     Lao Cai Province, Vietnam \\
                    \citet{Tracy2018} &                              MAXENT &                         Western North America \\
                \citet{Vacchiano2018} &                              MAXENT &                  Aosta Valley, Northern Italy \\
          \citet{Fernandez-Manso2019} &                              MAXENT &                               Valencia, Spain \\
            \citet{Ghorbanzadeh2019a} &                                 ANN &            Mazandaran Province, Northern Iran \\
            \citet{Ghorbanzadeh2019b} &                        ANN, SVM, RF &            Mazandaran Province, Northern Iran \\
                  \citet{Gigovic2019} &                             SVM, RF &                    Tara National Park, Serbia \\
                  \citet{Jaafari2019} &                             RF, SVM &                        Zagros Mountains, Iran \\
                 \citet{Jaafari2019a} &                             NFM, GA &                     Hyrcanian ecoregion, Iran \\
                      \citet{Kim2019} &                          MAXENT, RF &                                   South Korea \\
                      \citet{Lim2019} &                              MAXENT &                                   South Korea \\
                   \citet{Martin2019} &                              MAXENT &                               Northeast Spain \\
                 \citet{Mpakairi2019} &                              MAXENT &                         northwestern Zimbabwe \\
                 \citet{Quintano2019} &                              MAXENT &                             La Cabrera, Spain \\
                    \citet{Rihan2019} &                                  RF &                             Mongolian Plateau \\
                  \citet{Syphard2019} &                              MAXENT &                               California, USA \\
                    \citet{Zhang2019} &                   CNN, RF, SVM, ANN &                        Yunnan Province, China \\
\end{longtable}

\subsection*{S.3.4 Landscape controls on fire}

\begin{longtable}{p{4cm}p{4.5cm}p{6cm}}
\toprule
                    Citation &    ML methods used &                                  Study Region \\
\midrule
\endhead
\midrule
\multicolumn{3}{r}{{Continued on next page}} \\
\midrule
\endfoot

\bottomrule
\endlastfoot
        \citet{Amatulli2006} &                 DT &                      Gargano Peninsula, Italy \\
        \citet{Amatulli2007} &                 DT &                                Tuscany, Italy \\
       \citet{Archibald2009} &                 RF &                               Southern Africa \\
          \citet{Holden2009} &                 RF &         Gila National Forest, New Mexico, USA \\
              \citet{Li2009} &                ANN &                                         Japan \\
        \citet{Parisien2009} &        MAXENT, BRT &                                           USA \\
       \citet{Vasilakos2009} &                ANN &                         Lesvos Island, Greece \\
         \citet{Dlamini2010} &                 BN &                                     Swaziland \\
       \citet{Aldersley2011} &             DT, RF &                                        global \\
        \citet{Dimuccio2011} &                ANN &                              Central Portugal \\
      \citet{Sitanggang2011} &                 DT &      Rokan Hilir District, Sumatra, Indonesia \\
          \citet{Viedma2011} &            BRT, DT &           Guadalajara province, Central Spain \\
        \citet{Bisquert2012} &                ANN &                      Galicia, Northwest Spain \\
          \citet{Moritz2012} &             MAXENT &                                        Global \\
        \citet{Oliveira2012} &                 RF &                          Mediterranean Europe \\
        \citet{Parisien2012} &             MAXENT &                                   Western USA \\
          \citet{Renard2012} &             MAXENT &                          Western Ghats, India \\
         \citet{Syphard2012} &             MAXENT &                      Southern California, USA \\
      \citet{BarMassada2013} &         RF, MAXENT &                                 Michigan, USA \\
        \citet{Batllori2013} &             MAXENT &               Mediterranen ecosystems, Global \\
            \citet{Liu2013a} &                BRT &   Great Xing'an Mountains, Northeastern China \\
             \citet{Luo2013} &                 RF &                                        global \\
          \citet{Peters2013} &             MAXENT &                                 Northeast USA \\
      \citet{Sitanggang2013} &                 DT &      Rokan Hilir District, Sumatra, Indonesia \\
         \citet{Syphard2013} &             MAXENT &  South coast ecoregion, San Diego County, USA \\
          \citet{Arpaci2014} &         RF, MAXENT &                          Tyrol, European Alps \\
        \citet{Lydersen2014} &                 RF &        central Sierra Nevada, California, USA \\
         \citet{Maxwell2014} &                 RF &                                               \\
        \citet{Parisien2014} &                 DT &                                        Canada \\
      \citet{Rodrigues2014a} &       RF, BRT, SVM &                              Peninsular Spain \\
              \citet{Wu2014} &             RF, DT &                Great Xing'an Mountains, China \\
       \citet{Arganaraz2015} &                BRT &                             Central Argentina \\
            \citet{Chen2015} &             MAXENT &    Daxinganling Mountains, Northeastern China \\
        \citet{Chingono2015} &             MAXENT &                               Southern Africa \\
            \citet{Curt2015} &                BRT &                                 New Caledonia \\
           \citet{Duane2015} &             MAXENT &                              Catalonia, Spain \\
            \citet{Kane2015} &                 RF &                Sierra Nevada, California, USA \\
            \citet{Liu2015a} &                BRT &                                   Western USA \\
           \citet{Parks2015} &                BRT &                                   Western USA \\
          \citet{Viedma2015} &                BRT &           Guadalajara province, Central Spain \\
     \citet{Vijayakumar2015} &                 RF &                        central Quebec, Canada \\
              \citet{Wu2015} &            KNN, DT &                Great Xing'an Mountains, China \\
         \citet{Bashari2016} &                 BN &                        Isfahan province, Iran \\
      \citet{Coppoletta2016} &             RF, DT &       Northern Sierra Nevada, California, USA \\
            \citet{Curt2016} &                BRT &                           Southeastern France \\
       \citet{Fernandes2016} &                BRT &                                      Portugal \\
         \citet{Fonseca2016} &             MAXENT &                                        Brazil \\
        \citet{Goldarag2016} &                ANN &              Golestan province, Northern Iran \\
             \citet{Guo2016} &                 RF &          Daxing'an Mountains, Northeast China \\
            \citet{Guo2016a} &                 RF &                        Fujian province, China \\
 \citet{Miquelajauregui2016} &             RF, DT &                        central Quebec, Canada \\
       \citet{Pourtaghi2016} &            BRT, RF &              Golestan province, Northern Iran \\
           \citet{Satir2016} &                ANN &                    Upper Seyhan Basin, Turkey \\
         \citet{Syphard2016} &             MAXENT &  South coast ecoregion, San Diego county, USA \\
           \citet{Vilar2016} &             MAXENT &                          Madrid region, Spain \\
            \citet{Adab2017} &                ANN &                                Northeast Iran \\
             \citet{Cao2017} &            ANN, RF &                        Yunnan Province, China \\
           \citet{Davis2017} &             MAXENT &                        Pacific Northwest, USA \\
          \citet{Dwomoh2017} &                BRT &             Upper Guinean Region, West Africa \\
        \citet{Ebrahimy2017} &             MAXENT &                            Eastern Azerbaijan \\
          \citet{Forkel2017} &             RF, GA &                                        global \\
          \citet{Harris2017} &                 RF &                Sierra Nevada, California, USA \\
            \citet{Leys2017} &                 RF &                     Central Great Plains, USA \\
              \citet{Li2017} &             MAXENT &              Yunnan Province, Southwest China \\
        \citet{Lydersen2017} &                 RF &        central Sierra Nevada, California, USA \\
         \citet{Mostafa2017} &                SVM &              Golestan province, Northern Iran \\
          \citet{Nelson2017} &        DT, BRT, RF &                      British Columbia, Canada \\
          \citet{Peters2017} &             MAXENT &                                 Northeast USA \\
            \citet{Song2017} &                 RF &                             Hefei City, China \\
         \citet{TienBui2017} &  NFM, PSO, RF, SVM &                    Lam Dong province, Vietnam \\
          \citet{Valdez2017} &                 RF &                                      Honduras \\
           \citet{Young2017} &                BRT &                                   Alaska, USA \\
            \citet{Adab2018} &             MAXENT &                     Mazandaran province, Iran \\
            \citet{Fang2018} &                BRT &                Great Xing'an Mountains, China \\
            \citet{Hong2018} &        SVM, RF, GA &                    Southwest Jiangxi Province \\
         \citet{Jaafari2018} &         DT, DT, NB &                        Zagros Mountains, Iran \\
           \citet{Kahiu2018} &                BRT &                            sub Saharan Africa \\
          \citet{Masrur2018} &                 RF &                            circumpolar arctic \\
            \citet{Mayr2018} &    DT, RF, SVM, KM &                                       Namibia \\
           \citet{Parks2018} &                BRT &                                   Western USA \\
        \citet{Tehrany2018a} &        LB, SVM, RF &                     Lao Cai Province, Vietnam \\
           \citet{Tracy2018} &             MAXENT &                         Western North America \\
       \citet{Vacchiano2018} &             MAXENT &                  Aosta Valley, Northern Italy \\
            \citet{Ying2018} &                 RF &                                         China \\
          \citet{Clarke2019} &             MAXENT &                           Southeast Australia \\
 \citet{Fernandez-Manso2019} &             MAXENT &                               Valencia, Spain \\
          \citet{Forkel2019} &                 RF &                                        global \\
    \citet{GarciaLlamas2019} &                 RF &                               Northwest Spain \\
   \citet{Ghorbanzadeh2019a} &                ANN &            Mazandaran Province, Northern Iran \\
   \citet{Ghorbanzadeh2019b} &       ANN, SVM, RF &            Mazandaran Province, Northern Iran \\
         \citet{Gigovic2019} &            SVM, RF &                    Tara National Park, Serbia \\
         \citet{Jaafari2019} &            RF, SVM &                        Zagros Mountains, Iran \\
        \citet{Jaafari2019a} &            NFM, GA &                     Hyrcanian ecoregion, Iran \\
             \citet{Kim2019} &         MAXENT, RF &                                   South Korea \\
          \citet{Mansuy2019} &             MAXENT &                                 North America \\
          \citet{Molina2019} &             MAXENT &                     Andalusia, southern Spain \\
        \citet{Mpakairi2019} &             MAXENT &                         northwestern Zimbabwe \\
           \citet{Rihan2019} &                 RF &                             Mongolian Plateau \\
         \citet{Syphard2019} &             MAXENT &                               California, USA \\
           \citet{Young2019} &                BRT &                                   Alaska, USA \\
           \citet{Zhang2019} &  CNN, RF, SVM, ANN &                        Yunnan Province, China \\
\end{longtable}

\section*{S.4. Fire Behaviour Prediction}

\subsection*{S.4.1 Fire Spread and Growth}

\begin{longtable}{p{4cm}p{4.5cm}p{6cm}}
\toprule
                         Citation & ML methods used &                       Study Region \\
\midrule
\endhead
\midrule
\multicolumn{3}{r}{{Continued on next page}} \\
\midrule
\endfoot

\bottomrule
\endlastfoot
              \citet{Vakalis2004} &             ANN &              Attica region, Greece \\
             \citet{Abdalhaq2005} &              GA &                      not specified \\
            \citet{Rodriguez2008} &              GA &                   Catalonia, Spain \\
             \citet{Markuzon2009} &     RF, BN, KNN &                      Southwest USA \\
            \citet{Rodriguez2009} &              GA &                   Catalonia, Spain \\
           \citet{Cencerrado2012} &              GA &  Ashley National Forest, Utah, USA \\
               \citet{Denham2012} &              GA &                  Gestosa, Portugal \\
           \citet{Cencerrado2013} &              GA &                   Catalonia, Spain \\
                \citet{Kozik2013} &             ANN &                      not specified \\
                \citet{Artes2014} &              GA &                    Northeast Spain \\
           \citet{Cencerrado2014} &              GA &                   Catalonia, Spain \\
                \citet{Kozik2014} &             RNN &                      not specified \\
               \citet{Ascoli2015} &              GA &                    Southern Europe \\
           \citet{Chetehouna2015} &             ANN &                  Experimental data \\
                \citet{Artes2016} &              GA &         Northeast Catalonia, Spain \\
             \citet{Carrillo2016} &              GA &             Arkadia region, Greece \\
                \citet{Artes2017} &              GA &         Northeast Catalonia, Spain \\
          \citet{Subramanian2017} &              RL &           Northern Alberta, Canada \\
                \citet{Zheng2017} &             ANN &                        western USA \\
               \citet{Denham2018} &              GA &   Northern Patagonia Andean region \\
 \citet{GanapathiSubramanian2018} &              RL &           Northern Alberta, Canada \\
               \citet{Hodges2019} &             CNN &                    California, USA \\
              \citet{Khakzad2019} &              BN &                             Canada \\
                \citet{Radke2019} &             CNN &                                USA \\
\end{longtable}

\subsection*{S.4.2 Burned area and fire severity prediction}

\begin{longtable}{p{4cm}p{4.5cm}p{6cm}}
\toprule
                Citation &             ML methods used &                       Study Region \\
\midrule
\endhead
\midrule
\multicolumn{3}{r}{{Continued on next page}} \\
\midrule
\endfoot

\bottomrule
\endlastfoot
      \citet{Cortez2007} &            DT, RF, ANN, SVM &  Montesinho natural park, Portugal \\
          \citet{Yu2011} &                    SOM, ANN &  Montesinho natural park, Portugal \\
   \citet{Ozbayoglu2012} &                    ANN, SVM &                             Turkey \\
        \citet{Safi2013} &                         ANN &  Montesinho natural park, Portugal \\
 \citet{Zwirglmaier2013} &                          BN &                             Cyprus \\
      \citet{Shidik2014} &   ANN, NB, DT, RF, KNN, SVM &  Montesinho natural park, Portugal \\
         \citet{Xie2014} &                         SVM &         Guangzhou City area, China \\
      \citet{Alberg2015} &                          DT &  Montesinho natural park, Portugal \\
   \citet{Castelli2015a} &                          GA &  Montesinho natural park, Portugal \\
  \citet{Naganathan2016} &                SVM, DT, KNN &                                USA \\
      \citet{Storer2016} &                    ANN, PSO &  Montesinho natural park, Portugal \\
 \citet{Mitsopoulos2017} &                     BRT, RF &                             Greece \\
    \citet{AlJanabi2018} &                    ANN, SVM &  Montesinho natural park, Portugal \\
          \citet{Li2018} &                          DT &  Montesinho natural park, Portugal \\
     \citet{Toujani2018} &                    HMM, SOM &                  northwest Tunisia \\
        \citet{Zald2018} &                          RF &              southwest Oregon, USA \\
    \citet{Coffield2019} &       DT, RF, ANN, KNN, BRT &                        Alaska, USA \\
       \citet{Liang2019} &              ANN, RNN, LSTM &                    Alberta, Canada \\
         \citet{Xie2019} &  BRT, DNN, DT, SVM, ANN, RF &  Montesinho Natural Park, Portugal \\
\end{longtable}

\section*{S.5. Fire Effects}

\subsection*{S.5.1 Soil erosion and deposits}

\begin{longtable}{p{4cm}p{4.5cm}p{6cm}}
\toprule
             Citation & ML methods used &       Study Region \\
\midrule
\endhead
\midrule
\multicolumn{3}{r}{{Continued on next page}} \\
\midrule
\endfoot

\bottomrule
\endlastfoot
 \citet{Mallinis2009} &          DT, KM &             Greece \\
 \citet{Buckland2019} &             ANN &      Nebraska, USA \\
 \citet{Quintano2019} &          MAXENT &  La Cabrera, Spain \\
\end{longtable}

\subsection*{S.5.2 Smoke and particulate levels}

\begin{longtable}{p{4cm}p{4.5cm}p{6cm}}
\toprule
            Citation &             ML methods used &              Study Region \\
\midrule
\endhead
\midrule
\multicolumn{3}{r}{{Continued on next page}} \\
\midrule
\endfoot

\bottomrule
\endlastfoot
    \citet{Reid2015} &  RF, BAG, BRT, SVM, GP, KNN &           California, USA \\
 \citet{Lozhkin2016} &                         ANN &    Irkutsk Region, Russia \\
     \citet{Yao2018} &                          RF &  British Columbia, Canada \\
    \citet{Yao2018a} &                          RF &  British Columbia, Canada \\
 \citet{Fuentes2019} &                         ANN &           South Australia \\
  \citet{Watson2019} &      BRT, RF, SVM, KNN, ANN &  Northern California, USA \\
     \citet{Zou2019} &                     BRT, RF &    Pacific Northwest, USA \\
\end{longtable}

\subsection*{S.5.3 Post-fire regeneration and ecology}

\begin{longtable}{p{4cm}p{4.5cm}p{6cm}}
\toprule
                 Citation & ML methods used &                                 Study Region \\
\midrule
\endhead
\midrule
\multicolumn{3}{r}{{Continued on next page}} \\
\midrule
\endfoot

\bottomrule
\endlastfoot
    \citet{Johnstone2010} &             BRT &                                  Alaska, USA \\
     \citet{Thompson2010} &          RF, DT &  northwest California, southwest Oregon, USA \\
      \citet{Barrett2011} &              RF &                                       Alaska \\
        \citet{Perry2012} &             BRT &                                  New Zealand \\
       \citet{Reside2012} &          MAXENT &                                    Australia \\
     \citet{Sherrill2012} &             BRT &              Dinosaur National Monument, USA \\
          \citet{Cai2013} &             BRT &              Huzhong National Reserve, China \\
       \citet{Debouk2013} &             ANN &                             Catalonia, Spain \\
         \citet{Jung2013} &          GA, RF &             Central Siberian Plateau, Russia \\
       \citet{Chapin2014} &              RF &                                  Alaska, USA \\
         \citet{Liu2014a} &             BRT &               Huzhong Natural Reserve, China \\
          \citet{Han2015} &              RF &                       Yunnan province, China \\
   \citet{Hermosilla2015} &              RF &                         Saskatchewan, Canada \\
        \citet{Divya2016} &              NB &                                        India \\
 \citet{Martin-Alcon2016} &              RF &                             Catalonia, Spain \\
  \citet{Vijayakumar2016} &              RF &                               Quebec, Canada \\
      \citet{Fairman2017} &              RF &                          Victoria, Australia \\
          \citet{Luo2017} &              DT &                     Cangshan Mountain, China \\
    \citet{Papakosta2017} &              BN &                                       Cyprus \\
         \citet{Joao2018} &              RF &                            northern Portugal \\
         \citet{Poon2018} &             SVM &              San Bernardino, California, USA \\
       \citet{Cardil2019} &             BRT &                          southwestern Europe \\
    \citet{Magadzire2019} &          MAXENT &          Cape Floristic Region, South Africa \\
\end{longtable}

\subsection*{S.5.4 Socioeconomic effects}

\begin{longtable}{p{4cm}p{4.5cm}p{6cm}}
\toprule
            Citation & ML methods used &             Study Region \\
\midrule
\endhead
\midrule
\multicolumn{3}{r}{{Continued on next page}} \\
\midrule
\endfoot

\bottomrule
\endlastfoot
 \citet{Hradsky2017} &              BN &  Otway Ranges, Australia \\
\end{longtable}

\section*{S.6. Fire Management}

\subsection*{S.6.1 Planning and policy}

\begin{longtable}{p{4cm}p{4.5cm}p{6cm}}
\toprule
             Citation & ML methods used &                            Study Region \\
\midrule
\endhead
\midrule
\multicolumn{3}{r}{{Continued on next page}} \\
\midrule
\endfoot

\bottomrule
\endlastfoot
   \citet{Penman2011} &              BN &        Wollemi National Park, Australia \\
      \citet{Bao2015} &              GA &  Longdong Forest Park, Guangzhou, China \\
 \citet{Ruffault2015} &                 &                                         \\
  \citet{Bradley2016} &              RF &                             western USA \\
 \citet{McGregor2016} &             MDP &                           not specified \\
 \citet{McGregor2017} &              RF &  Deschutes National Forest, Oregon, USA \\
\end{longtable}

\subsection*{S.6.2 Fuel treatment}

\begin{longtable}{p{4cm}p{4.5cm}p{6cm}}
\toprule
            Citation & ML methods used &              Study Region \\
\midrule
\endhead
\midrule
\multicolumn{3}{r}{{Continued on next page}} \\
\midrule
\endfoot

\bottomrule
\endlastfoot
 \citet{Penman2014a} &              BN &    Southeastern Australia \\
    \citet{Arca2015} &              GA &  Southern Sardinia, Italy \\
   \citet{Lauer2017} &              RL &          southwest Oregon \\
\end{longtable}

\subsection*{S.6.3 Wildfire preparedness and response}

\begin{longtable}{p{4cm}p{4.5cm}p{6cm}}
\toprule
                       Citation & ML methods used &                          Study Region \\
\midrule
\endhead
\midrule
\multicolumn{3}{r}{{Continued on next page}} \\
\midrule
\endfoot

\bottomrule
\endlastfoot
       \citet{Homchaudhuri2010} &              GA &                         not specified \\
 \citet{Costafreda-Aumedes2015} &             ANN &                                 Spain \\
             \citet{Penman2015} &              BN &     Sydney Basin Bioregion, Australia \\
            \citet{OConnor2017} &     BRT, MAXENT &  Southern Idaho, Northern Nevada, USA \\
            \citet{Julian2018a} &          RL, DL &                            simulation \\
          \citet{Rodrigues2019} &              BN &                      Catalonia, Spain \\
\end{longtable}

\subsection*{S.6.4 Social factors}

\begin{longtable}{p{4cm}p{4.5cm}p{6cm}}
\toprule
            Citation & ML methods used & Study Region \\
\midrule
\endhead
\midrule
\multicolumn{3}{r}{{Continued on next page}} \\
\midrule
\endfoot

\bottomrule
\endlastfoot
 \citet{Delgado2018} &              BN &        Spain \\
\end{longtable}